# Transform Quantization for CNN Compression

Sean I. Young, *Member, IEEE*, Wang Zhe, *Member, IEEE*,
David Taubman, *Fellow, IEEE*, and Bernd Girod, *Fellow, IEEE*

**Abstract**—In this paper, we compress convolutional neural network (CNN) weights post-training via transform quantization. Previous CNN quantization techniques tend to ignore the joint statistics of weights and activations, producing sub-optimal CNN performance at a given quantization bit-rate, or consider their joint statistics during training only and do not facilitate efficient compression of already trained CNN models. We optimally transform (decorrelate) and quantize the weights post-training using a rate–distortion framework to improve compression at any given quantization bit-rate. Transform quantization unifies quantization and dimensionality reduction (decorrelation) techniques in a single framework to facilitate low bit-rate compression of CNNs and efficient inference in the transform domain. We first introduce a theory of rate and distortion for CNN quantization, and pose optimum quantization as a rate–distortion optimization problem. We then show that this problem can be solved using optimal bit-depth allocation following decorrelation by the optimal End-to-end Learned Transform (ELT) we derive in this paper. Experiments demonstrate that transform quantization advances the state of the art in CNN compression in both retrained and non-retrained quantization scenarios. In particular, we find that transform quantization with retraining is able to compress CNN models such as AlexNet, ResNet and DenseNet to very low bit-rates (1–2 bits).

✦

## 1 INTRODUCTION

CONVOLUTIONAL neural networks (CNNs) have become a universal framework for solving a number of problems in pattern analysis and computer vision, from classification [1], segmentation [2] and style transfer [3, 4] of images to the synthesis of images [5]. While CNNs outperform traditional non-learning-based methods in many vision problems, they can involve tens or hundreds of millions of parameters, and their sheer bulk renders deploying CNNs on mobile devices difficult. CNN weights no longer fit in the cache at the same time, and they must be loaded from the off-chip memory as required, which can be very energy-consuming [6]. Even the amount of off-chip memory can be limited on these devices so the storage of weights is itself a challenge. Moreover, the number of computations performed on these CNN weights may hinder the use of CNNs for certain interactive and real-time applications. One way to reduce the computational and the storage burden of CNNs is to compress and simplify the representation of their numerous weight parameters.

Recent CNN compression techniques may be categorized as weight pruning [7–21], weight quantization [20–46], and dimensionality reduction of weight matrices [47–56]. Out of these, weight quantization has shown particular usefulness for compressing CNNs to binary or tertiary representations [28–30], at various rate–accuracy trade-offs [36, 37] and even post-training [38] without any labeled data. However, while current quantization methods yield good compression, they tend to ignore the joint statistics of weights and activations and produce sub-optimum network performance at a given quantization bit-rate, or consider their joint statistics during training only [45], and do not facilitate efficient compression of already-trained networks.

For post-training compression, optimum decorrelation of the weights coupled with optimum bit-depth allocations can significantly improve compression especially at the low bit-rates similar to vector quantization [22–26], but without the increase in the encoding complexity from the irregularity of quantization cells. In practice, scalar quantization following a decorrelating transform is preferred to vector quantization as suggested by the success of image compression methods based on e.g. the Discrete Cosine Transform (DCT) [57] and the Discrete Wavelet Transform (DWT) [58]. The latest video coding standard VVC [59] can even select the best transform from DCT-II, -VIII and DST-VII, Trellis-coding the resulting transform coefficients to further improve compression.

In this work, we propose to quantize CNN weights post-training using transform quantization. We analyze the CNN compression problem from a rate–distortion viewpoint, and show that a particular decorrelating transform coupled with optimum bit-depths can solve this problem. Transforming a convolutional or a fully-connecting layer enables dimension reduction similar to PCA, while quantizing the insignificant transform kernels to zero at low rates emulates the behavior of pruning methods. Therefore, transform quantization uses all three ingredients of CNN compression—dimensionality reduction of weights, quantization, and pruning—and also enables mixed-precision inference on specialized hardware if different bit-depths are assigned to channels. In Fig. 1, we illustrate the two components (transform and quantization)

- *S. I. Young and B. Girod are with the Information Systems Laboratory (ISL), Department of Electrical Engineering, Stanford University, Stanford CA 94305, USA. E-mail: sean0@stanford.edu, bgirod@stanford.edu.*
- *W. Zhe was with the Information Systems Laboratory (ISL), Department of Electrical Engineering, Stanford University. He is now with the Institute for Infocomm Research, A\*STAR, Singapore.
Email: wang_zhe@i2r.a-star.edu.sg*
- *D. Taubman is with the School of Electrical Engineering and Telecommunications, University of New South Wales, Sydney, NSW, Australia.
E-mail: d.taubman@unsw.edu.au*

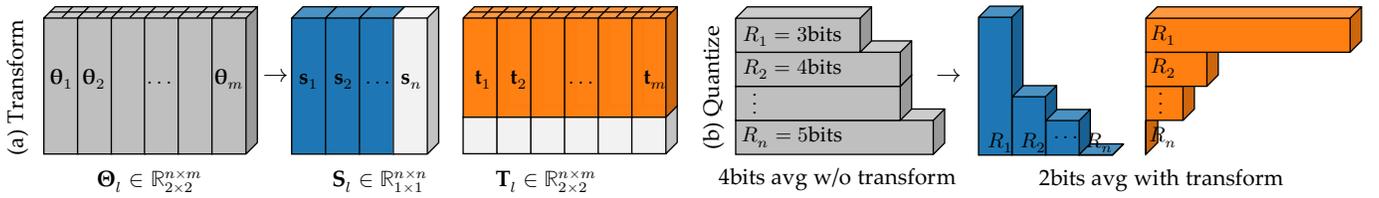

Fig. 1. Transform quantization of CNN layers. Given weight matrices $\boldsymbol{\Theta}_1, \boldsymbol{\Theta}_2, \ldots, \boldsymbol{\Theta}_L$ of an *L*-layer CNN, we represent each one as $\boldsymbol{\Theta}_l = \mathbf{S}_l \mathbf{T}_l$ (a) and quantize both the kernel matrix $\mathbf{T}_l$ and the basis $\mathbf{S}_l$ optimally (b). In (b), the bar lengths illustrate the bit-depths needed to quantize $\boldsymbol{\Theta}_l$ directly (gray bars), or $\mathbf{S}_l \mathbf{T}_l$ in the transform domain (blue and orange bars) for the same performance. Elements corresponding to zero bit-depth assignments ($R_k = 0$) are indicated as white blocks in (a).

of transform quantization. A $h \times w$ convolution layer (gray) is transformed to a pair of basis (blue) and transform domain (orange) convolutional layers (a), then the pair quantized to rate–distortion optimum bit-depths (b). In the absence of quantization, composing the mappings of this pair of layers recovers (or synthesizes) the original one. Signal processing literature uses the term "transform coding" more commonly than "transform quantization". However, we adopt the latter term in this work to stress our focus on quantization rather than (entropy) coding, which we address in our sequel.

Our main contributions are as follows. First, we propose transform quantization for compressing CNN weights—we are the first to consider quantization of transformed weights and basis, and optimize both post-training. Second, we give a theory of rate and distortion for CNN quantization, based on which transform coding gains can be computed. We then derive an end-to-end-learned transform that maximizes the gains. Third, we advance the state of the art in compression of CNNs, in both retrained and non-retrained scenarios, for image classification—AlexNet [1], ResNets [60], DenseNets [61], and for low-level vision tasks, DRUNet (denoising) [62] and EDSR (super-resolution) [63].

## 2 RELATED WORK

CNN compression methods based on pruning [7–20] nullify kernels [7–13], channels [14–17], or even individual weights [18–20] that are insignificant according to such criteria as a norm [7–9] or some objective function [18]. However, while a few methods [20] compress the pruned weights further via quantization and variable-length (Huffman) coding, optimal quantization of pruned networks generally remains an open problem. Moreover, the pruning criteria are often chosen in a heuristic manner agnostic to the actual accuracy losses. By contrast, our transform-quantization framework quantizes to zero (i.e. prunes) kernels which are deemed insignificant according to bit-depth–accuracy optimality criteria.

Contemporaneous with the pruning-based techniques for CNN compression, some researchers proposed to compress CNNs first by spatially decorrelating their weights with the DCT, followed by vector quantization [24] or hashing [26] of the decorrelated weights to further reduce the redundancies across the transformed kernels. These transform approaches are, however, applicable to the compression of convolution kernels only, computationally complex due to the clustering involved [24], or cache-inefficient due to using hashtables as the main data structure [26]. Moreover, some CNN models contain large numbers of $1 \times 1$ convolutional kernels, where the benefit of spatial decorrelation of kernels via the DCT is less significant. In contrast with the spatial-transform-based approaches, our quantization framework transforms across (rather than within) kernels to provide the compression and efficiency gains even for smaller convolution kernels.

In parallel with the DCT techniques above, other authors sought to reduce the dimensionality of CNN layers via PCA and related techniques [47–56]. Denton et al. [47] apply the SVD across kernels in convolution layers, or across the rows and columns of weight matrices for fully-connecting layers to project input or output to low-rank subspaces, providing computational savings during the inference stage. Zhang et al. [51] extend this method by using the generalized SVD to take into account the presence of ReLU activations between successive layers. This work was extended in turn by Kim et al. [52], who apply the SVD on all four axes of convolutional tensors. Li et al. [49] extend the work of [51] by partitioning the kernels into subsets, then using the SVD on each subset of kernels separately. However, PCA-based compression is not concerned with quantization, so optimal quantization of the projected network remains an open problem. Also, PCA methods determine the projection dimensions heuristically using decorrelated weight variances [51, 52], by minimizing a least-squares criterion [50] or using dimension assignment rules [54], none of which reflect the actual loss of prediction accuracy due to the projections. By contrast, our framework determines the quantization bit-depths and their associated projection dimensions to minimize the performance loss at a given quantization bit-rate (average bit-depth).

Quantization methods (both vector and scalar) for CNNs [20–46] have developed alongside pruning methods. Vector quantization of weights [22–26] involves $k$-means [22–24] or a "hashing trick" [25, 26] and may additionally involve the DCT [24, 26] and residual quantization [22]. Uniform scalar quantization of weights to one bit (binary representation) is proposed by Hubara et al. [31, 41], and to two bits (ternary) by Zhu et al. [30]. LQ-Nets [37] learn the optimal quantizers during training at a specified bit-depth. Non-uniform scalar quantization was proposed by Tung and Mori [21], whereas Zhou et al. [36] demonstrate that we can quantize networks post-training, provided that weights can be refined by some additional training after quantizing (we refer to this process as re-training). Recent methods [38, 43] propose to quantize networks in a training-free manner, based on a per-channel bit-depth allocation paradigm [38], or by equalizing weights and correcting biases [43]. We extend per-channel bit-depth allocation and re-training to the transform domain.

Designing lightweight CNNs such as MobileNet [64] and SqueezeNet [65] is another form of model compression. One can relate the depthwise convolution layers of [64] to perfect diagonalization of weight matrices using the SVD, while the reduced-input-channel $3 \times 3$ convolution [65] relates closely to the projection of input spaces to their lower-dimensional sub-spaces, both during training. However, discovering the optimum lightweight CNN models can be challenging since architecture design decisions must be made prior to training the network. Optimal quantization of the resulting network also remains a problem. Transform quantization is based on the same decorrelating (diagonalizing) principles, inducing a low-dimensional projection of weights at low quantization bit-rates. However, our framework has the advantage that it jointly addresses optimal quantization, post-training.

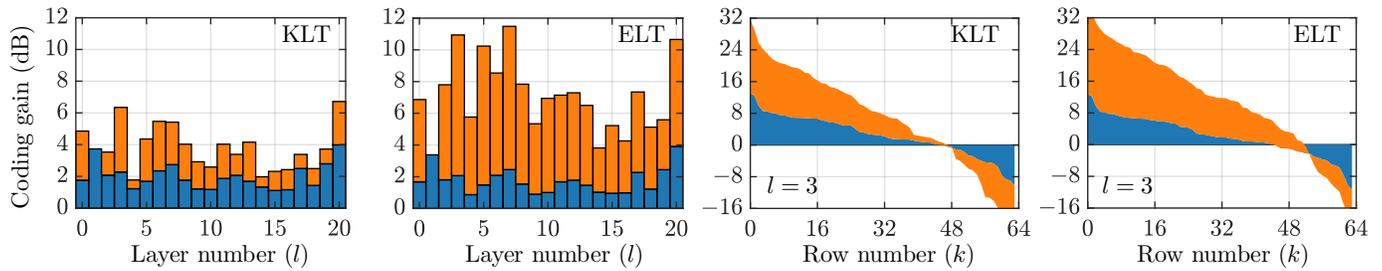

Fig. 3. Coding gains due to the KLT and the ELT applied onto the rows of weight matrices $\boldsymbol{\Theta}_l$. Convolution and fully-connected layers of ResNet-18 exhibit KLT coding gains of 1–7 dB and ELT ones of 1–12 dB (left plots), where the blue and the orange bars indicate the gain components due to the decorrelation of weights and gradients, respectively. A coding gain of $G$ produces a rate-saving of $\frac{1}{2}\log_2 G$. The per-matrix coding gains can be broken down further into per-column coding-gains shown in the right sub-plots for layers 3 of ResNet-18 (right).

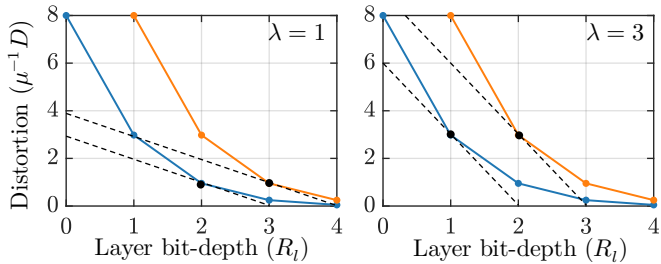

Fig. 2. Rate–distortion optimal layer bit-depth assignment. For a given rate–distortion trade-off $\lambda$, we sweep the bit-depth–distortion curve of each layer (blue and orange lines) to find the bit-depth values $R_l$ where the slope is equal to $-\lambda$ (black dots).

Oktay et al. [45] and Huang et al. [66] propose to impose kernel decorrelation at training time, rather than decorrelate them post hoc. Although such strategies can be beneficial for reducing model redundancy, they increase training time if a high degree of orthogonality is needed [66], do not facilitate compression of already-trained models, and leaves open the optimal choice of decorrelating transform [45] as well as the problem of optimally quantizing the decorrelated kernels. A related class of methods known as natural gradient descent [67–72] propose to accelerate the training of CNNs via local normalization of the loss landscape before taking descent. In contrast with natural gradient descent where transforms are used to whiten weight gradient, our framework decorrelates (but not whiten) the weights themselves prior to quantizing to improve the quantizer's "packing".

Gao et al. [73] study the limits of CNN compression from a rate–distortion perspective, reinterpreting compression as a rate-allocation problem similarly to this work. Their study however does not consider the use of a transform as part of quantization. Section 4 shows that applying a decorrelating transform before quantizing is key to good compression.

## 3 NOTATION AND REMARKS

Let us set our notation by defining a vector space over CNN input, output and weights, equipped with operations based on multi-input-multi-output (MIMO) convolutions [74]. We denote by $\mathbf{x} \in \mathbb{R}^m_{a \times b}$ an $m$-channel signal whose elements are matrices of size $a \times b$, that is, $x_k \in \mathbb{R}^{a \times b}$ for $k = 1, \ldots, m$. We denote by $\boldsymbol{\Theta} \in \mathbb{R}^{n \times m}_{a \times b}$ a convolutional layer with $m$ input and $n$ output channels whose convolution kernels $\theta_{kj} \in \mathbb{R}^{a \times b}$ for $k = 1, \ldots, n$ and $j = 1, \ldots, m$. One is now ready to equip the two spaces $\mathbb{R}^m_{a \times b}$ and $\mathbb{R}^{n \times m}_{a \times b}$ with the following vector space operations.

**Definition 1: Inner product.** The inner product of two given $m$-channel signals $\mathbf{x}, \mathbf{y} \in \mathbb{R}^m_{a \times b}$ can be defined as

$$\langle \mathbf{x}, \mathbf{y} \rangle = \langle x_1, y_1 \rangle_F + \cdots + \langle x_m, y_m \rangle_F \in \mathbb{R}, \quad (1)$$

in which $\langle x, y \rangle_F$ denotes the Frobenius inner product of the matrix-elements $x, y \in \mathbb{R}^{a \times b}$.

**Definition 2: Euclidean norm.** The Euclidean norm of given $m$-channel signal $\mathbf{x} \in \mathbb{R}^m_{a \times b}$ can be defined using (1) as

$$\|\mathbf{x}\|_2 = \sqrt{\langle \mathbf{x}, \mathbf{x} \rangle} = \sqrt{\|x_1\|_F^2 + \cdots + \|x_m\|_F^2} \in \mathbb{R}_+, \quad (2)$$

in which $\|x\|_F$ denotes the Frobenius norm of $x \in \mathbb{R}^{a \times b}$.

**Definition 3: Layer–layer product.** The layer-layer product $\mathbf{Z} = \mathbf{XY}$ of $\mathbf{X} \in \mathbb{R}^{o \times n}_{a \times b}$ and $\mathbf{Y} \in \mathbb{R}^{n \times m}_{c \times d}$ is defined as

$$\mathbf{Z}_{kj} = (x_{k1} \star y_{1j} + \cdots + x_{kn} \star y_{nj}) \in \mathbb{R}^{(|a-c|+1) \times (|b-d|+1)} \quad (3)$$

for $k = 1, \ldots, o$ and $j = 1, \ldots, m$, where $\star$ represent valid (as opposed to full or same) 2D convolution[1]. Observe that each $x \star y = \langle x, y \rangle_F$ if $x$ and $y$ have the same dimensions.

In contrast with Definitions 1–2, Definition 3 can involve elements from two (possibly different) vector spaces. We can analogously derive the definitions of layer–signal and outer-products from Definition 3. The transpose $\boldsymbol{\Theta}^t \in \mathbb{R}^{m \times n}_{a \times b}$ of a layer $\boldsymbol{\Theta} \in \mathbb{R}^{n \times m}_{a \times b}$ is defined via $(\boldsymbol{\Theta}^t)_{jk} = \boldsymbol{\Theta}_{kj}$ in analogy with that of a conventional matrix. With the definitions above, we can express the mapping of an $m$-channel image $\mathbf{x} \in \mathbb{R}^m_{c \times d}$ by a convolution layer $\boldsymbol{\Theta} \in \mathbb{R}^{n \times m}_{a \times b}$ with $m$ inputs and $n$ outputs compactly as $\mathbf{x} \mapsto \boldsymbol{\Theta}\mathbf{x}$. Notice that because filters form a ring (but not a field) with respect to convolution, classical matrix decompositions like the LU and the QR cannot be applied to convolution layer $\boldsymbol{\Theta}$. However, a linear transform $\boldsymbol{\Theta} = \mathbf{ST}$ is still well-defined. In this work, we essentially transform all convolutional mappings of the form $\mathbf{x} \mapsto \boldsymbol{\Theta} \mathbf{x}$ to $\mathbf{x} \mapsto \mathbf{STx}$ and quantize $\mathbf{S}$ and $\mathbf{T}$. See Fig. 1 (left), where $\boldsymbol{\Theta}, \mathbf{T} \in \mathbb{R}^{n \times m}_{2 \times 2}$ depict $2 \times 2$ convolution layers and $\mathbf{S} \in \mathbb{R}^{n \times n}_{1 \times 1}$, a basis layer, which is equivalent to a $1 \times 1$ convolution layer.

---

1. Valid convolution $z = x \star y \in \mathbb{R}^{a-b+1}$ of $x \in \mathbb{R}^a$ and $y \in \mathbb{R}^b$ (when $a \geq b$) is given by $z_j = \sum_{i=1}^m x_{j+i-1} y_i$ for $j = 1, \ldots, a-b+1$.

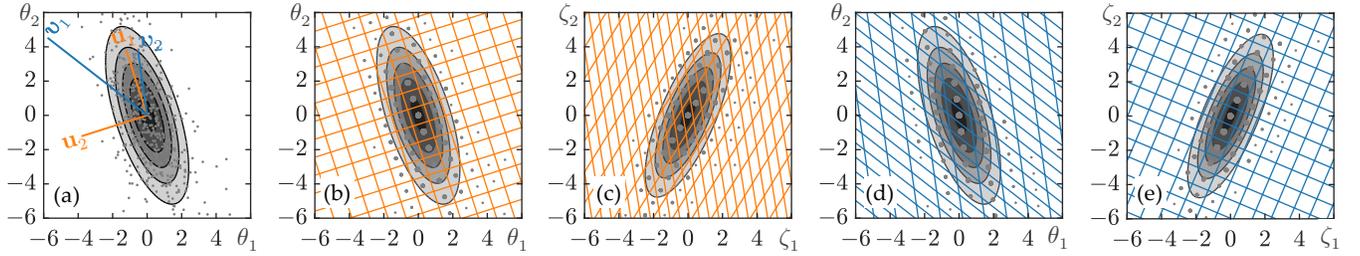

Fig. 4. The KLT and the ELT. Elements $\theta_1$ and $\theta_2$ of columns $\boldsymbol{\theta}$ of weight matrix $\boldsymbol{\Theta} \in \mathbb{R}^{2 \times 512}$, shown as dots in (a), are correlated with an underlying Gaussian distribution depicted by the level sets. Elements $\gamma_1$ and $\gamma_2$ of $\boldsymbol{\gamma} = \partial y / \partial \boldsymbol{\theta}$ (not shown) are correlated with covariance matrix $\mathbf{C}_{\gamma\gamma}$. The KLT $\mathbf{U}^t$ of $\boldsymbol{\theta}$ is an orthogonal projection $\mathbf{t} = \mathbf{U}^t \boldsymbol{\theta}$ of $\boldsymbol{\theta}$ onto $\mathbf{u}_1$ and $\mathbf{u}_2$, and induces quantization cells that are square in the domain of $\boldsymbol{\theta}$ (b) but parallelograms in the domain of $\boldsymbol{\zeta} = \mathbf{C}_{\gamma\gamma}^{1/2} \boldsymbol{\theta}$ (c). On the other hand, the ELT $\mathbf{U}^t$ is a biorthogonal projection of $\boldsymbol{\theta}$ onto $\mathbf{u}_1$ and $\mathbf{u}_2$, and induces cells that are parallelograms in the domain of $\boldsymbol{\theta}$ (d) but square in the domain of $\boldsymbol{\zeta}$ (e), minimizing distortion in the output domain.

## 4 QUANTIZATION OF CNNs

This work addresses the problem of compressing the weight matrices of a CNN. We can express the end-to-end mapping of an $L$-layer feed-forward CNN as

$$\mathbf{y} = f(\mathbf{x}|\boldsymbol{\Theta}_1, \ldots, \boldsymbol{\Theta}_L) = f_L(\cdots f_2(\boldsymbol{\Theta}_2 \, f_1(\boldsymbol{\Theta}_1 \mathbf{x}))), \quad (4)$$

in which $\mathbf{x} \in \mathbb{R}^M_{a \times b}$ and $\mathbf{y} \in \mathbb{R}^N_{c \times d}$ are respectively the network input and output, and $\boldsymbol{\Theta}_{l=1,\ldots,L}$ are the $L$ convolutional and fully-connecting weight matrices that parameterize $f$. Non-linearities between convolutions, such as pooling, activation (ReLU or tanh), and normalization have been absorbed into functions $f_l$ together with biases. Although bias parameters are also learnable such that they too must be quantized and stored, they are sufficiently few relative to weights, and their impact on the quantized size of CNNs is negligible. When a classification CNN is being quantized, output $\mathbf{y}$ is assumed to be the logits, that is, result prior to softmax activation.

To give a concrete example, AlexNet [1] parameterizes $f$ using $L = 8$ convolutional and fully-connecting weights, the input $\mathbf{x} \in \mathbb{R}^3_{224 \times 224}$ represents a 3-color image with $224 \times 224$ pixels, and output $\mathbf{y} \in \mathbb{R}^{1000}_{1 \times 1}$ unnormalized log-probabilities of membership of input $\mathbf{x}$ across the 1000 predefined classes (tench, ..., toilet paper). Convolution layer $\boldsymbol{\Theta}_1 \in \mathbb{R}^{64 \times 3}_{11 \times 11}$ has $64 \times 3$ convolutional kernels of size $11 \times 11$, while the last fully-connected layer can be denoted $\boldsymbol{\Theta}_8 \in \mathbb{R}^{1000 \times 4096}_{1 \times 1}$.

### 4.1 Layer-wise Quantization

The elements of weights $\boldsymbol{\Theta}_l$ are continuously-valued so that they necessitate quantization for efficient communication or storage. In the simplest case, we can use uniform quantizers on weights directly. Quantizing $\boldsymbol{\Theta}_l$ at a bit-depth of $R_l$ gives us $\boldsymbol{\Theta}_l^q = q(\boldsymbol{\Theta}_l, R_l, \Delta_l(R_l))$, where $q(\theta, R, \Delta) = 0$ if $R = 0$ and

$$q(\theta, R, \Delta) = \Delta \, \text{clip}(\text{round}(\Delta^{-1}\theta), -2^{R-1}, 2^{R-1} - 1) \quad (5)$$

otherwise, where $R_l$ is a whole number of bits and $\Delta_l(R_l)$ is the optimum quantization step-size for a given $R_l$. Denoting by $\hat{\mathbf{y}} = f(\mathbf{x}|\boldsymbol{\Theta}^q_{l=1,\ldots,L})$ the output produced by the quantized network, our CNN compression problem can be seen as one of minimizing $\mathbb{E}\|\hat{\mathbf{y}} - \mathbf{y}\|^2_2$ across the distribution of $\mathbf{y}$, subject to upper bound $R_{\max}$ on the average quantization bit-depth (quantization bit-rate) required to represent $\boldsymbol{\Theta}^q_1, \ldots, \boldsymbol{\Theta}^q_L$:

$$\begin{aligned} \text{minimize} \quad & D(R_1, \ldots, R_L) = \mathbb{E}_{\mathbf{x} \sim P(\mathbf{x})} \|\hat{\mathbf{y}} - \mathbf{y}\|^2_2 \\ \text{subject to} \quad & R(R_1, \ldots, R_L) = \sum_{l=1}^{L} \mu_l R_l \leq R_{\max} \end{aligned} \quad (6)$$

[44], where the expectation is taken over the distribution of a training dataset, and $\mu_l$ denote the fraction of weights in $\boldsymbol{\Theta}^q_l$ over the total. We use distortion with mean-squared error as a proxy for the true accuracy measure of interest such as the top-1 classification error, treating the CNN $f$ as a regression model (in particular, we find the top-1 accuracy metric to be monotonic in mean-squared error). If we relax the inequality constraint of (6), we obtain

$$\text{minimize} \quad J = D(R_1, \ldots, R_L) + \lambda R(R_1, \ldots, R_L), \quad (7)$$

in which $\lambda$ decides the trade-off between rate and distortion optimization criteria. Differentiating the objective of (7) with respect to bit-depths $R_l$, we obtain the bit-depth–distortion (or rate–distortion) optimality conditions

$$\lambda = -\frac{1}{\mu_1} \frac{\partial D}{\partial R_1} = -\frac{1}{\mu_2} \frac{\partial D}{\partial R_2} = \cdots = -\frac{1}{\mu_L} \frac{\partial D}{\partial R_L} \quad (8)$$

cf. [75] so assuming one is allocating an infinitesimal bit, the optimal trade-off is reached when the decrease in the output distortion from this infinitesimal bit is equal for all layers.

Optimality conditions (8) suggest a discrete approach for solving problem (7). We first generate a bit-depth–distortion curve $\mu_l^{-1} D(\ldots, R_l, \ldots)$ for each layer $l$, keeping the others unquantized. Given some $\lambda$, the Lagrangian-optimum value of $R_l$ is one which minimizes $\mu_l^{-1} D(R_l) + \lambda R_l$. We show this optimality condition geometrically in Fig. 2. This procedure is discrete and requires no derivatives of $J$ to solve (7).

### 4.2 Transform Quantization

Certain rows, columns or elements of the convolutional and fully-connected weight matrices $\boldsymbol{\Theta}_l$ can have a larger impact on output distortion when quantized. Allocating bit-depths equally to all elements of $\boldsymbol{\Theta}_l$ may thus produce sub-optimal rate-accuracy performance. More significantly, there may be statistical correlation among the elements of $\boldsymbol{\Theta}_l$. We can use a decorrelating transform $\mathbf{U}_l^t$ on $\boldsymbol{\Theta}_l$ to obtain $\mathbf{T}_l = \mathbf{U}_l^t \boldsymbol{\Theta}_l$, and allocate bit-depths to transform elements $\mathbf{T}_l$ to maximize the quantizer's performance. For simplicity, we assume that our transforms are column ones ($\mathbf{T}_l = \mathbf{U}_l^t \boldsymbol{\Theta}_l$) as opposed to row ones ($\mathbf{T}_l = \boldsymbol{\Theta}_l \mathbf{U}_l$). To treat quantization statistically, it is also convenient to regard the columns of $\boldsymbol{\Theta}_l$ as realizations of an underlying random source $\boldsymbol{\theta}_l$, and columns of $\mathbf{T}_l$ as those of random source $\mathbf{t}_l$ and assume each of the random sources is jointly Gaussian. For brevity, we may omit the subscript $l$ of marices $\boldsymbol{\Theta}_l$, $\mathbf{T}_l$ and sources $\boldsymbol{\theta}_l$, $\mathbf{t}_l$, and write $\boldsymbol{\Theta}$, $\mathbf{T}$ and $\boldsymbol{\theta}$, $\mathbf{t}$. We first derive theoretical results for minimum distortion when the random source $\boldsymbol{\theta}$ is quantized without a transform, and bit-depths assigned to the individual elements of $\boldsymbol{\theta}$.

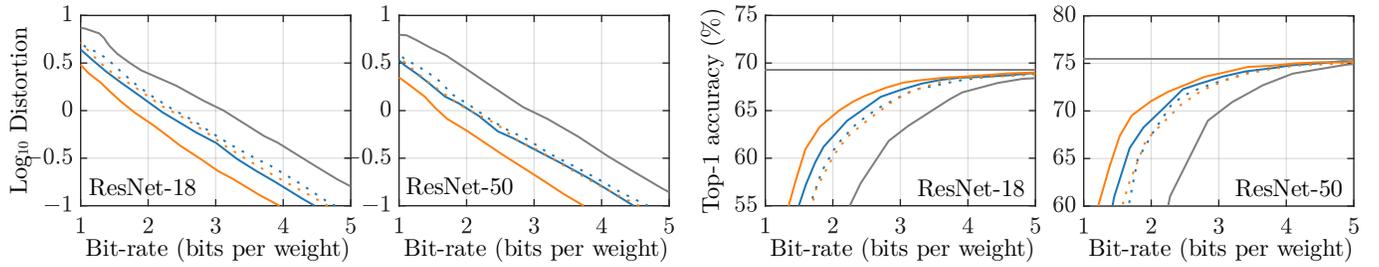

Fig. 6. Log10 output distortion (left plots) and top-1 accuracy (right plots) of Resnet-18 and 50 quantized with row (solid curves) and column (dotted curves) variants of the KLT (blue curves) and the ELT (orange curves). All transforms have better rate–accuracy trade-offs than no transform (gray curves), providing a rate saving of 1 bit across all rates. Bit-rates of the KLT and the ELT include bits spent on the transform matrices.

**Theorem 1: Minimum distortion without transform.** Given a jointly-Gaussian source $\boldsymbol{\theta} \in \mathbb{R}^n_{c \times d}$, the minimum distortion $D = \frac{1}{n}\mathbb{E}\|\mathbf{y} - \hat{\mathbf{y}}\|_2^2$ due to the quantization of $\boldsymbol{\theta}$ at a sufficiently high quantization bit-rate (average bit-depth) $R$ is given by

$$D_{\text{pcm}} = (\prod_{k=1}^n (\mathbf{C}_{\gamma\gamma})_{kk}(\mathbf{C}_{\theta\theta})_{kk})^{1/n}\epsilon^2 2^{-2R}, \quad (9)$$

in which $\mathbf{C}_{\theta\theta}$ and $\mathbf{C}_{\gamma\gamma}$ represent the cross-channel covariance matrix of $\boldsymbol{\theta} \in \mathbb{R}^n_{c \times d}$ and $\boldsymbol{\Gamma} = (\partial \mathbf{y}/\partial \boldsymbol{\theta}) \in \mathbb{R}^{n \times N}_{c \times d}$ respectively, $R$ is the average quantization bit-depth for the $n$ elements of $\boldsymbol{\theta}$ and $\epsilon^2$ is a constant which depends on the quantization and coding scheme used [76]. We give the proof in Appendix A.

Theorem 1 essentially states the output distortion due to quantizing the elements of $\boldsymbol{\theta}$ with optimal bit-depths for an average of $R$ bits is identical to distortion due to quantizing a Gaussian source of variance $(\prod_{k=1}^n (\mathbf{C}_{\gamma\gamma})_{kk}(\mathbf{C}_{\theta\theta})_{kk})^{1/n}$ at $R$ bits. The exponential decay $2^{-2R}$ comes from the halving of the quantization step-size with each additional bit, reducing output distortion by a factor of four. We now improve upon the rate-distortion function (9) by transforming the elements of $\boldsymbol{\theta}$ prior to quantization. Suppose a transform $\mathbf{U}^t \in \mathbb{R}^{n \times n}_{1 \times 1}$ is first applied on $\boldsymbol{\theta}$ to produce coefficients $\mathbf{t} = \mathbf{U}^t\boldsymbol{\theta}$, and $\mathbf{t}$ then quantized. Quantization bit-depths $R_k$ are allocated now to the elements of transform source $\mathbf{t}$. Theorem 2 now derives the minimum output distortion for transform quantization.

**Theorem 2: Minimum distortion with transform.** Given the same jointly Gaussian source $\boldsymbol{\theta} \in \mathbb{R}^n_{c \times d}$, the minimum output distortion $D = \frac{1}{n}\mathbb{E}\|\mathbf{y} - \hat{\mathbf{y}}\|_2^2$ caused by the quantization of the transform coefficients $\mathbf{t} = \mathbf{U}^t\boldsymbol{\theta}$ at sufficiently high rates $R$ is

$$D_{\text{tc}} = (\prod_{k=1}^n (\mathbf{U}^{-1}\mathbf{C}_{\gamma\gamma}\mathbf{U}^{-t})_{kk}(\mathbf{U}^t\mathbf{C}_{\theta\theta}\mathbf{U})_{kk})^{1/n}\epsilon^2 2^{-2R} \quad (10)$$

in which $\mathbf{C}_{\gamma\gamma}$ and $\mathbf{C}_{\theta\theta}$ are defined identically as before, and $R$ is now the average quantization bit-depth allocated to the $n$ elements of $\mathbf{t}$. We provide the proof in Appendix A.

The ratio between the two distortions (9) and (10),

$$G = \frac{(\prod_{k=1}^n (\mathbf{C}_{\gamma\gamma})_{kk})^{1/n}}{(\prod_{k=1}^n (\mathbf{U}^{-1}\mathbf{C}_{\gamma\gamma}\mathbf{U}^{-t})_{kk})^{1/n}} \frac{(\prod_{k=1}^n (\mathbf{C}_{\theta\theta})_{kk})^{1/n}}{(\prod_{k=1}^n (\mathbf{U}^t\mathbf{C}_{\theta\theta}\mathbf{U})_{kk})^{1/n}} \quad (11)$$

is known as the coding gain [77] of transform $\mathbf{U}^t$ (higher is better). We expect quantization of $\mathbf{t} = \mathbf{U}^t\boldsymbol{\theta}$ to produce a bit-rate saving of $\frac{1}{2}\log_2 G$ bits over that of the original $\boldsymbol{\theta}$ source for a given output distortion. The classical Karhunen–Loève Transform (KLT) of $\boldsymbol{\theta}$ (i.e. the matrix of eigenvectors of $\mathbf{C}_{\theta\theta}$) is however sub-optimal with respect to $G$ since it maximizes only the second gain term of (11). The optimal transform for CNN quantization is now derived.

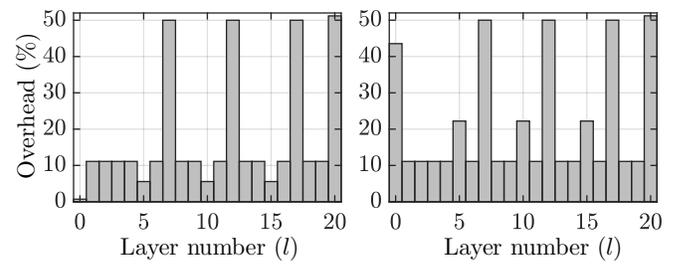

Fig. 5. ResNet-18 transform overheads as percentages of the number of elements in $\boldsymbol{\Theta}_l$. Row (left) and column (right) transform overheads shown. Layers 7, 12, 17, and 20 are fully connected.

**Theorem 3: Optimum transforms for CNNs.** The transform $\mathbf{U}^t$ that minimizes output distortion $D = \frac{1}{n}\mathbb{E}\|\mathbf{y} - \hat{\mathbf{y}}\|_2^2$ caused by the quantization of $\mathbf{t} = \mathbf{U}^t\boldsymbol{\theta}$, diagonalizes the matrix pair $(\mathbf{C}_{\theta\theta}, \mathbf{C}_{\gamma\gamma}^{-1})$. That is,

$$\mathbf{U}^t\mathbf{C}_{\theta\theta}\mathbf{U} = \boldsymbol{\Lambda} \quad \text{and} \quad \mathbf{U}^t\mathbf{C}_{\gamma\gamma}^{-1}\mathbf{U} = \mathbf{I}, \quad (12)$$

in which $\mathbf{C}_{\theta\theta}$ and $\mathbf{C}_{\gamma\gamma}$ represent the cross-channel covariance matrix of $\boldsymbol{\theta}$ and $\boldsymbol{\Gamma}$ respectively as previously, and $\boldsymbol{\Lambda}$ denotes the nonnegative diagonal matrix of generalized eigenvalues of matrix pair $(\mathbf{C}_{\theta\theta}, \mathbf{C}_{\gamma\gamma}^{-1})$. In general, $\mathbf{U}^{-1} \neq \mathbf{U}^t$ since $\mathbf{U}$ is not necessarily orthogonal. The proof is given in Appendix A.

We refer to the optimal transform $\mathbf{U}^t$ from Theorem 3 as the End-to-end Learned Transform (ELT) of $\boldsymbol{\theta}$. The ELTs are so-called since they simultaneously decorrelate the weights $\boldsymbol{\theta}$ and the partial derivatives $\partial \mathbf{y}/\partial \boldsymbol{\theta}$, which are derived from the end-to-end mapping of $\mathbf{x}$ to output $\mathbf{y} = f(\mathbf{x}|\boldsymbol{\Theta}_1, \ldots, \boldsymbol{\Theta}_L)$ across all values of $\mathbf{x}$. In practice, we can compute $\partial \mathbf{y}/\partial \boldsymbol{\theta}$ by back-propagation using a small validation set. In Fig. 3, we compare the coding gains of the KLT and the ELT across the layers of ResNet-18. We see that the ELT produces a further 4 dB gain over the KLT. Both transforms were applied onto the rows of the weight matrices.

Fig. 4 (a–e) illustrate the difference between the KLT and the ELT. Consider a simple multi-layer perceptron

$$y = f(\mathbf{x}|\boldsymbol{\Theta}_1, \boldsymbol{\Theta}_2, \ldots, \boldsymbol{\Theta}_L), \quad (13)$$

in which $\mathbf{x} \in \mathbb{R}^N$, and $y \in \mathbb{R}$. Suppose now that elements $\theta$ of

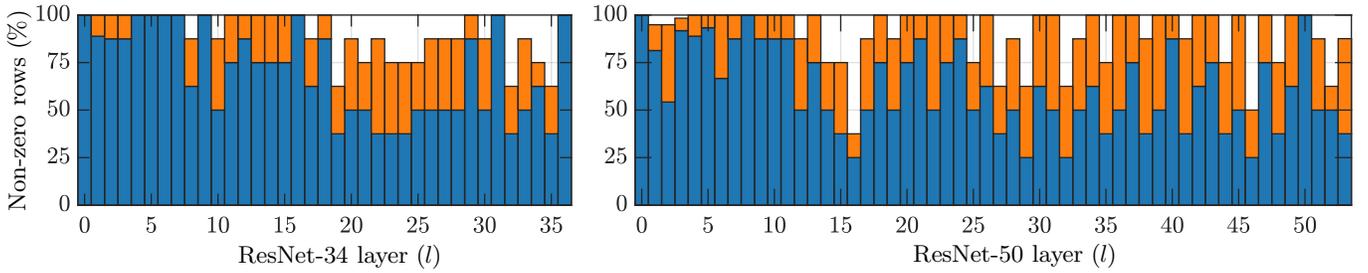

Fig. 7. Percentages of non-zero rows of transformed weight matrices $\mathbf{T}_l$ at bit-rates $R = 1.1$ and 2.1 bits for ResNet-34 (left), and $R = 1.1$ and 3.1 bits for ResNet-50 (right). Percentages of non-zero rows at lower and higher bit-rates visualized as blue and blue + orange bars, respectively. See Fig. 9 for retrained and non-retrained classification accuracies at these bit-rates.

---

**Algorithm 1.** Optimal Bit-depth Allocation

1  **Input:** $\mathbf{Z} \in \mathbb{R}_{a \times b}^{m \times n}, \lambda, D_k, \Delta_k, k = 1, \ldots, m$
2  **Output:** $\Delta_k^{\text{opt}}, R_k^{\text{opt}}, k = 1, \ldots, m$
3  **for** $k = 1, \ldots, m$ **do**
4  $\quad J = \infty, R_k = 0$
5  $\quad$ **for** $R = 0, \ldots, M$ **do**
6  $\quad\quad$ **if** $J > D_k(R) + \lambda mnabR$ **then**
7  $\quad\quad\quad J = D_k(R) + \lambda mnabR$
8  $\quad\quad\quad R_k^{\text{opt}} = R$
9  $\quad\quad\quad \Delta_k^{\text{opt}} = \Delta_k(R)$
10 $\quad\quad$ **end if**
11 $\quad$ **end for**
12 **end for**

---

columns $\boldsymbol{\theta}_{j=1,\ldots,512} \in \mathbb{R}^2$ of some $\boldsymbol{\Theta}_l \in \mathbb{R}^{2 \times 512}$, shown as dots in (a), are correlated, as are the elements $\gamma$ of the derivatives $\boldsymbol{\gamma}_j = \partial y / \partial \boldsymbol{\theta}_j \in \mathbb{R}^2$ for all $j$. We can visualize the mapping by the KLT $\mathbf{U}^t$ of $\boldsymbol{\theta}$ as an orthogonal projection of $\boldsymbol{\theta}$ onto its two principal axes $\mathbf{u}_1$ and $\mathbf{u}_2$. Quantizing the projected elements $\mathbf{t} = \mathbf{U}^t \boldsymbol{\theta}$ maps $\mathbf{t}$ to the centers of square cells (b). Noting that distortion can be written as $\|\mathbf{C}_{\gamma\gamma}^{1/2} (\boldsymbol{\theta} - \boldsymbol{\theta}^q)\|_2^2$, the square cells become parallelograms in the domain of $\boldsymbol{\zeta} = \mathbf{C}_{\gamma\gamma}^{1/2} \boldsymbol{\theta}$ (c) and incur larger distortion than square cells of the same area. On the other hand, the ELT $\mathbf{U}^t$ of $\boldsymbol{\theta}$ is a biorthogonal projection of $\boldsymbol{\theta}$ onto nonorthogonal vectors $\boldsymbol{v}_1$ and $\boldsymbol{v}_2$ (a). Quantization of $\mathbf{t} = \mathbf{U}^t \boldsymbol{\theta}$ now maps $\mathbf{t}$ to the centers of parallelogram cells (d). These parallelogram cells map to squares in the domain of $\boldsymbol{\zeta}$ (e), and minimizes distortion $\|\mathbf{C}_{\gamma\gamma}^{1/2} (\boldsymbol{\theta} - \boldsymbol{\theta}^q)\|_2^2$ as a result.

Fig. 6 (left) shows that the ELT (orange curves) produces lower output distortion than the KLT (blue curves) for row- (solid) and column- (dotted) variants of the two transforms in case where no retraining is performed. The classification accuracy associated with each distortion is graphed in Fig. 6 (right). We see that the KLT still provides good classification performance while being significantly easier to construct as it does not require the derivatives ($\partial \mathbf{y} / \partial \boldsymbol{\theta}$). Row transforms perform better than column ones, suggesting weights have a larger correlation within rows than in columns. We relate the KLT and the ELT to SVD and GSVD in Appendix B.

### 4.3 Quantizing Transform Bases

Although the coding gain expression in (11) provides a good indication of the expected rate savings due to transform, we need to additionally count the bits spent on quantizing and storing the inverse transform $\mathbf{S} = \mathbf{U}^{-t}$, required at inference time, together with layer $\mathbf{T}$ to reconstruct the original layer $\boldsymbol{\Theta}$. The overhead amount depends on whether we transform the columns ($\mathbf{T} = \mathbf{U}^t \boldsymbol{\Theta}$) or the rows ($\mathbf{T} = \boldsymbol{\Theta} \mathbf{U}$) of the weights $\boldsymbol{\Theta} \in \mathbb{R}_{a \times b}^{n \times m}$. Again, suppose the former, and $n < mab$. In this case, overheads due to the transform are the $n^2$ elements of matrix $\mathbf{S}$. If $n \geq mab$, however, only the first $mab$ (non-zero) rows of $\mathbf{T}$ need be stored, together with their corresponding $mab$ columns of $\mathbf{S}$. As a result, the total number of overheads is $\min(n^2, (mab)^2)$ relative to quantizing $\boldsymbol{\Theta}$ directly without transform. We can analyze the row-transform case similarly.

Fig. 5 plots the transform-quantization overheads for the layers of ResNet-18 as percentages of the number of weights in each one. We consider the overhead for both the row- and the column-transformed cases. In both cases, the transforms add overheads of roughly 10% of the number of weights. To allocate bit-depths optimally to $\mathbf{T}$ and $\mathbf{S}$ at some given rate–distortion trade-off $\lambda$, we must satisfy optimality conditions (8) on the rows of $\mathbf{T}$, and columns of $\mathbf{S}$. Bit-depth allocation procedure for $\mathbf{T}$ and $\mathbf{S}$ is given in Algorithm 1. Input $\mathbf{Z}$ can be transform domain $\mathbf{T}$, or (the transpose of) the basis $\mathbf{S}$.

At lower bit-rates, less significant rows of $\mathbf{T} \in \mathbb{R}_{a \times b}^{n \times m}$ may be quantized so heavily that only some first $k < n$ rows of $\mathbf{T}$ are non-zero (column-ELT and column-KLT sort the rows of $\mathbf{T}$ in a non-ascending order of variances). In this case, we can evaluate the mapping $\mathbf{x} \mapsto \boldsymbol{\Theta} \mathbf{x}$ efficiently in the transformed domain via $\mathbf{x} \mapsto \mathbf{S}_{1:k}(\mathbf{T}_{1:k}\mathbf{x})$, where matrices $\mathbf{S}_{1:k} \in \mathbb{R}^{n \times k}_{1 \times 1}$ and $\mathbf{T}_{1:k} \in \mathbb{R}^{k \times m}_{a \times b}$ are the first $k$ columns and the rows of $\mathbf{S}$ and $\mathbf{T}$ respectively. Since the first mapping $\mathbf{x} \mapsto \mathbf{T}_{1:k}\mathbf{x}$ entails larger $a \times b$ convolutions whereas the second one $\mathbf{z} \mapsto \mathbf{S}_{1:k}\mathbf{z}$, $1 \times 1$ ones, we achieve a large inference speed-up when $k$ is small.

One can express this inference speed-up as the ratio

$$A = (1^2 nk + abmk)/(abmn), \quad (14)$$

that is, the ratio of the number of multiplications in the map $\mathbf{x} \mapsto \boldsymbol{\Theta} \mathbf{x}$ to that in $\mathbf{x} \mapsto \mathbf{S}_{1:k}(\mathbf{T}_{1:k}\mathbf{x})$. Typically, $abm \gg 1^2 n$, so the acceleration $A \approx k/n$ may be understood as the fraction of non-zero rows of $\boldsymbol{\Theta}$. In Fig. 7, we graph the percentage of non-zero rows of ResNet-18 and ResNet-34 weight matrices $\boldsymbol{\Theta}$ at 1–2 bits, illustrating that significant acceleration can be obtained. Averaging (14) across all $L$ layers weighted by the sizes of the input activations gives us the overall acceleration in the number of floating-point operations (FLOPs).

## 5 OPTIMIZING QUANTIZATION

We now discuss optimal scalar quantization of the elements of $\mathbf{t} = \mathbf{U}^t \boldsymbol{\theta}$. While vector [77] and dead-zone [78] quantizers can further improve upon the rate–accuracy performance of transform quantization, we opt to use simple uniform scalar quantization to accelerate inference using integer arithmetic directly on quantization indices.

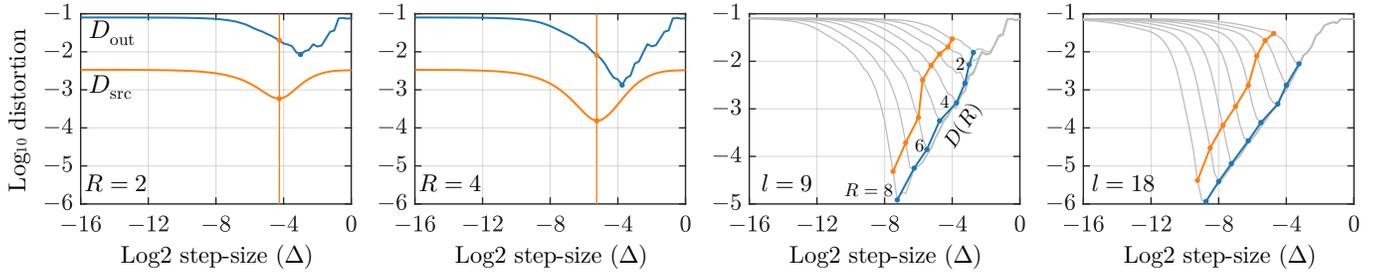

Fig. 8. Weight and output distortions against quantization step-size $\Delta$ at bit-depths of $R = 2$ and $4$ (left, shown for the first row-KLT element of layer 9 in ResNet-18). The step-size that minimizes the weight distortion $D_{\text{src}}$ (orange lines) does not necessarily minimize the output distortion $D_{\text{out}}$ and incurs a 1–2-bit-rate loss compared with using the optimal quantization step-size for the output (right, shown for layers 9 and 18 of ResNet-18).

TABLE 1
Quantization times of different CNNs on Intel Xeon 6132 @ 2.60GHz + Nvidia Quadro RTX 8000

| Model | Weights | Layers | Blocks | Steps | Maxbits | Perf. | Cost |
|---|---|---|---|---|---|---|---|
| AlexNet | 62,378k | 8 | 8 | 8 | 16 | 7.0ms | 1.6h |
| ResNet-18 | 11,679k | 21 | 8 | 8 | 16 | 7.0ms | 4.2h |
| ResNet-34 | 21,780k | 37 | 8 | 8 | 16 | 7.1ms | 7.5h |
| ResNet-50 | 25,503k | 54 | 8 | 8 | 16 | 7.2ms | 10.9h |
| DenseNet-121 | 7,894k | 121 | 4 | 8 | 16 | 7.6ms | 13.1h |

## 5.1 Finding Optimal Step-sizes

To find the optimum quantization step-size $\Delta$ for a random source $\mathbf{t}$ at a given bit-depth $R$, we can evaluate over a range of $\Delta$, output distortion $D_{\text{out}} = \mathbb{E}\|\mathbf{y} - \hat{\mathbf{y}}\|_2^2$ due to quantizing source $\mathbf{t}^q(\Delta)$, and find a value of $\Delta$ that minimizes $D_{\text{out}}$ (the number of quantization levels is fixed at $2^R$). Step-sizes that minimize the distortion $D_{\text{src}} = \mathbb{E}\|\mathbf{t} - \mathbf{t}^q\|_2^2$ of the source itself deviate significantly from the one which minimizes output distortion $D_{\text{out}}$. Fig. 8 (left) shows this for the KLT source $\mathbf{t}$ of the ninth layer of ResNet-18. This discrepancy in step-sizes results in a 1–2 bit-rate-loss (Fig. 8, right). The characteristic shape-$V$ curves in the right plots are attributed to an initial decrease in overload distortion as step-size $\Delta$ increases, then an increase in granular distortion past optimum $\Delta$—see [77] for further discussion on granular and overload distortions.

Banner et al. [38] derived an expression to relate optimal quantization step-sizes to the variance of the random source in the Laplace-distributed case. However, Fig. 8 suggests the quantization step-size should be optimized over the output distortion explicitly using a grid search. This explicit search also obviates the need to fit a parametric distribution on the underlying source, and is also applicable to quantizers with dead-zones [78], and non-uniform cells [79, 80]. Algorithm 2 lists the procedure for optimal quantization step-size search on $\mathbf{T}$. Step-size searches on $\mathbf{S}$ is similar, but involves $\mathbf{S}^q$ and $\mathbf{T}$ instead of $\mathbf{T}^q$ and $\mathbf{S}$ (being $\mathbf{S}_k$ now the $k$th column of $\mathbf{S}$).

## 5.2 Reducing Quantization Complexity

To maximize compression, one could allocate an individual bit-depth to each row of $\mathbf{T}$ and each column of $\mathbf{S}$. This can be time-consuming for layers having many channels since one needs to first construct a bit-depth–distortion curve for each row of $\mathbf{T}$, and column of $\mathbf{S}$. Each bit-depth–distortion curve requires in turn computing the output distortion for a large number of (bit-depth, step-size) pairs. To save time required to construct these curves, we therefore partition $\mathbf{T}$ (resp. $\mathbf{S}$) into $B$ blocks of rows (resp. columns), and allocate instead a

**Algorithm 2.** Optimal Step-size Search

1: **Input:** $\mathbf{S} \in \mathbb{R}^{n \times n}_{1 \times 1}$, $\mathbf{T} \in \mathbb{R}^{n \times m}_{c \times d}$
2: **Output:** $\Delta_k, D_k, k = 1, \ldots, n$
3: **for** $k = 1, \ldots, n$ **do**
4:     **for** $R = 1, \ldots, M$ **do**
5:         $D_k(R) = \infty, \Delta_k(R) = 0$
6:         **for** $\Delta = 2^0, 2^1, 2^2, \ldots$ **do**
7:             $\mathbf{T}^q = \mathbf{T}$ /* Initialize with unquantized weights */
8:             $\mathbf{T}^q_k = q(\mathbf{T}_k, R, \Delta)$ /* $\mathbf{T}_k$ is the $k$th row of $\mathbf{T}$ */
9:             $\hat{\mathbf{y}} = f(\mathbf{x}|\boldsymbol{\Theta}_1, \ldots, \mathbf{S}\mathbf{T}^q, \ldots, \boldsymbol{\Theta}_L)$
10:            **if** $D_k(R) > \mathbb{E}\|\mathbf{y} - \hat{\mathbf{y}}\|_2^2$ **then**
11:                 $D_k(R) = \mathbb{E}\|\mathbf{y} - \hat{\mathbf{y}}\|_2^2$
12:                 $\Delta_k(R) = \Delta$
13:            **end if**
14:         **end for**
15:     **end for**
16: **end for**

single bit-depth to each block. This has very little impact on rate–distortion performance if the number of blocks is large enough—the output variance of elements of $\mathbf{T}$ (resp. $\mathbf{S}$) is monotonic-decreasing across the rows (resp. columns), and the output variances roughly the same within each block. If the blocks can be allocated a maximum bit-depth of $M$, then time spent on constructing bit-depth–distortion curves of all layers is proportional to $BSIMPL$, where $S$ is the number of quantization step-sizes searched, $I$, number of images used to generate the bit-depth–distortion curves, $P$, performance of a given CNN in seconds, and $L$ the number of layers. The typical parameters and associated times for optimizing bit-depth are shown in Table 1 for several networks.

## 5.3 Fine-tuning Quantized Weights

Whereas fine-tuning of the quantized networks is not always possible due to lack of suitable training data at compression time, we demonstrate that optionally fine-tuning transform-quantized CNNs can restore accuracies close to the original ones. To re-train the transform-quantized networks, we use the straight-through estimator (STE) approach of Hubara et al. [31, 41] to quantize the weights during the forward-pass according to the assigned bit-depths, and let their gradients pass through with no change during the backward pass. We plot in Fig. 9 the bit-rate–accuracy curves of ResNets on the ImageNet [81] validation set with and without retraining on the training set. The classification accuracies can be restored close to the original ones at 2 bits, and almost losslessly at 3 bits. All models were retrained for 3 epochs maximum with SGD, and the actual number of iterations optimized using a grid search to achieve minimum validation error. Note that transform bases **S** are also retrained so there is no difference between the KLT and the ELT cases after retraining.

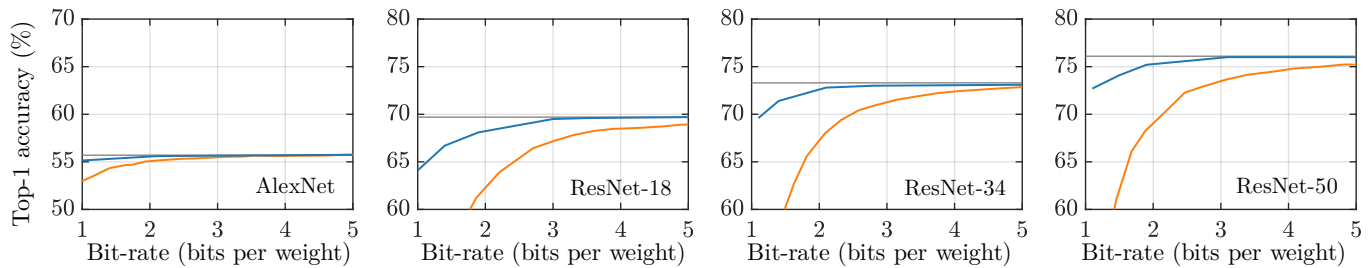

Fig. 9. Rate–accuracy curves of transform-quantized AlexNet and ResNets on the ImageNet validation set, with retraining (blue lines), and without retraining (orange lines). Unquantized baselines shown as gray lines. All CNNs retrained on the training set for 3 epochs maximum. Shown for the row-KLT case. Retrained row-ELT results are similar (no distinction exists between the KLT and the ELT after retraining).

TABLE 2
Classification accuracies of CNN models compressed using transform quantization and other methods.

| Methods | Train epochs | Comp ratio | Wgt/Act bit-rate | Top-1 (%) | Top-5 (%) |
|---|---|---|---|---|---|
| **ResNet-18** | | | | | |
| Full-precision | – | – | 32/32 | 69.7 (+0.0) | 89.2 (+0.0) |
| BWN [28] | 16 | 32.0× | 1/32 | 60.8 (–8.9) | 83.0 (–6.2) |
| TWN [29] | 50 | 16.0× | 2/32 | 65.3 (–4.4) | 86.2 (–3.0) |
| TTQ [30] | 160 | 16.0× | 2/32 | 66.6 (–3.1) | 87.2 (–2.0) |
| INQ [36] | 8 | 16.0× | 2/32 | 66.0 (–3.7) | 87.1 (–2.1) |
| INQ [36] | 8 | 10.7× | 3/32 | 68.1 (–1.6) | 88.4 (–0.8) |
| INQ [36] | 8 | 8.0× | 4/32 | 68.9 (–0.8) | 89.0 (–0.2) |
| INQ [36] | 8 | 6.4× | 5/32 | 69.0 (–0.7) | 89.1 (–0.1) |
| LQ-Nets [37] | 120 | 16.0× | 2/32 | 68.0 (–1.7) | 88.0 (–1.2) |
| LQ-Nets [37] | 120 | 10.7× | 3/32 | 69.3 (–0.4) | 88.8 (–0.4) |
| LQ-Nets [37] | 120 | 8.0× | 4/32 | 70.0 (+0.3) | 89.1 (–0.1) |
| DSQ [40] | 90 | 32.0× | 1/32 | 63.7 (–6.0) | – |
| DFQ [43] | 0 | 5.3× | 6/32 | 66.3 (–3.4) | – |
| Ours (row-ELT) | 0 | 10.7× | 3.0/32 | 67.9 (–1.8) | 88.1 (–1.1) |
| Ours (row-ELT) | 0 | 8.2× | 3.9/32 | 68.6 (–1.1) | 88.5 (–0.7) |
| Ours (row-KLT) | 0 | 10.7× | 3.0/32 | 67.2 (–2.5) | 87.8 (–1.4) |
| Ours (row-KLT) | 0 | 8.2× | 3.9/32 | 68.5 (–1.2) | 88.4 (–0.8) |
| Ours + retrain | 3 | 32.0× | 1.0/32 | 64.5 (–5.2) | 86.4 (–2.8) |
| Ours + retrain | 3 | 22.9× | 1.4/32 | 67.2 (–2.5) | 87.8 (–1.4) |
| Ours + retrain | 3 | 16.8× | 1.9/32 | 68.2 (–1.5) | 88.4 (–0.8) |
| Ours + retrain | 3 | 10.7× | 3.0/32 | 69.8 (+0.1) | 89.3 (+0.1) |
| **ResNet-34** | | | | | |
| Full-precision | – | – | 32/32 | 73.3 (+0.0) | 91.3 (+0.0) |
| L-DNQ [39] | – | 10.7× | 3/32 | 44.0 (–29.) | 72.9 (–18.) |
| HAQ [46] | 100 | 15.2× | 2.1/32 | 71.5 (–1.8) | 90.1 (–1.2) |
| HAQ [46] | 100 | 10.7× | 3.0/32 | 73.2 (–0.1) | 91.3 (+0.0) |
| Ours (row-ELT) | 0 | 11.9× | 2.7/32 | 71.5 (–1.8) | 90.3 (–1.0) |
| Ours (row-ELT) | 0 | 8.4× | 3.8/32 | 72.6 (–0.7) | 90.9 (–0.4) |
| Ours (row-KLT) | 0 | 11.4× | 2.8/32 | 71.0 (–2.3) | 90.2 (–1.1) |
| Ours (row-KLT) | 0 | 8.4× | 4.0/32 | 72.4 (–0.9) | 90.9 (–0.4) |
| Ours + retrain | 3 | 29.1× | 1.1/32 | 69.6 (–3.7) | 88.9 (–2.4) |
| Ours + retrain | 3 | 22.9× | 1.4/32 | 71.4 (–1.9) | 89.7 (–1.6) |
| Ours + retrain | 3 | 15.2× | 2.1/32 | 72.8 (–0.5) | 90.6 (–0.7) |
| Ours + retrain | 3 | 11.9× | 2.8/32 | 73.0 (–0.3) | 90.8 (–0.5) |
| **ResNet-50** | | | | | |
| Full-precision | – | – | 32/32 | 76.0 (+0.0) | 93.0 (+0.0) |
| INQ [36] | 8 | 6.4× | 5/32 | 74.8 (–1.2) | 92.4 (–0.6) |
| LQ-Nets [37] | 120 | 16.0× | 2/32 | 75.1 (–0.9) | 92.3 (–0.7) |
| LQ-Nets [37] | 120 | 8.0× | 4/32 | 76.4 (+0.4) | 93.1 (+0.1) |
| ABGD [23] | 9 | 32.0× | 1.0/32 | 68.2 (–7.8) | – |
| ABGD [23] | 9 | 18.8× | 1.7/32 | 73.8 (–2.2) | – |
| HAQ [46] | 100 | 10.6× | 3.0/32 | 75.3 (–0.7) | 92.5 (–0.5) |
| HAQ [46] | 100 | 8.0× | 4.0/32 | 76.1 (+0.1) | 92.9 (–0.1) |
| Ours (row-ELT) | 0 | 11.4× | 2.8/32 | 73.6 (–2.4) | 91.7 (–1.3) |
| Ours (row-ELT) | 0 | 8.4× | 3.8/32 | 74.7 (–1.1) | 92.4 (–0.6) |
| Ours (row-KLT) | 0 | 10.3× | 3.1/32 | 73.6 (–2.4) | 91.7 (–1.3) |
| Ours (row-KLT) | 0 | 7.8× | 4.1/32 | 74.8 (–1.2) | 92.3 (–0.7) |
| Ours + retrain | 3 | 29.1× | 1.1/32 | 72.9 (–3.1) | 91.4 (–1.6) |
| Ours + retrain | 3 | 21.3× | 1.5/32 | 74.2 (–1.8) | 92.2 (–0.8) |
| Ours + retrain | 3 | 16.8× | 1.9/32 | 75.2 (–0.8) | 92.6 (–0.4) |
| Ours + retrain | 3 | 11.4× | 3.1/32 | 76.0 (+0.0) | 93.0 (+0.0) |
| **AlexNet** | | | | | |
| Full-precision | – | – | 32/32 | 55.7 (+0.0) | 78.6 (+0.0) |
| BWN [28] | 58 | 32.0× | 1/32 | 56.8 (+1.1) | 79.4 (+0.8) |
| DoReFa [42] | 200 | 32.0× | 1/32 | 53.9 (–1.8) | 76.3 (–2.3) |
| TWN [29] | 50 | 16.0× | 2/32 | 54.5 (–1.2) | 76.8 (–1.8) |
| TTQ [30] | 160 | 16.0× | 2/32 | 57.5 (+1.8) | 79.7 (+1.1) |
| INQ [36] | 8 | 6.4× | 5/32 | 57.4 (+1.7) | 80.5 (+1.9) |
| LQ-Nets [37] | 120 | 16.0× | 2/32 | 60.5 (+4.8) | 82.7 (+4.1) |
| Ours (row-ELT) | 0 | 32.0× | 1.0/32 | 53.4 (–2.3) | 77.1 (–1.5) |
| Ours (row-ELT) | 0 | 16.0× | 2.0/32 | 55.1 (–0.6) | 78.2 (–0.4) |
| Ours (row-KLT) | 0 | 32.0× | 1.0/32 | 53.1 (–2.6) | 77.0 (–1.6) |
| Ours (row-KLT) | 0 | 16.0× | 2.0/32 | 55.0 (–0.7) | 78.2 (–0.4) |
| Ours + retrain | 3 | 64.0× | 0.5/32 | 53.0 (–2.7) | 77.0 (–1.6) |
| Ours + retrain | 3 | 32.0× | 1.0/32 | 55.3 (–0.4) | 78.5 (–0.1) |
| **DenseNet-121** | | | | | |
| Full-precision | – | – | 32/32 | 75.0 (+0.0) | 92.3 (+0.0) |
| Ours (row-KLT) | 0 | 8.0× | 4.0/32 | 72.6 (–2.4) | 91.1 (–1.2) |
| Ours (row-KLT) | 0 | 6.4× | 5.0/32 | 73.7 (–1.3) | 91.7 (–0.6) |

## 6 EXPERIMENTAL RESULTS

To examine the rate–distortion and rate–accuracy behaviors of transform-quantized networks, we compress a number of popular pretrained image classification networks—AlexNet [1], ResNets [60], and DenseNets [61]—as well as pretrained image superresolution network EDSR [63], in both retrained and non-retrained scenarios. We show transform-quantized CNNs advance the state of the art in both scenarios. For fair comparisons, we compare with other results that do not use entropy coding, e.g. Huffman [20] or arithmetic [45]. We can always apply standard entropy (lossless) coding to any of the quantized CNNs for a further ~30% reduction in bit-rates.

**Image classification accuracy.** Table 2 shows the accuracies of transform-quantized AlexNet, ResNets and DenseNets on ImageNet [81] (pretrained PyTorch versions) at 1–3 bits. The accuracies are evaluated using the ImageNet2012 validation set consisting of 50k images. Our CNNs are quantized using row-ELT and row-KLT. For reference, we include the results of DFQ [43], LQ-Nets [37], INQ [36], ABGD [23] DoReFaNet [42] and HAQ [46]. We see that our non-retrained ResNet-18 results are significantly better than those for the DFQ (data-free quantization) method [43]. In the case where retraining is performed, row-KLT ResNets have a considerably higher accuracy than HAQ at low bits (ResNet-34, 50), and slightly more accurate overall than LQ-Nets (ResNet-18, 50) but with 3 epochs of refining. LQ-Nets also requires the bit-depths to be determined prior to training the model while we are able to quantize and refine the models at arbitrary bit-rates after training. Note that our pretrained AlexNet model is smaller and has a lower unquantized baseline than the others. Bold numbers indicate results on the rate–top-1 frontier.

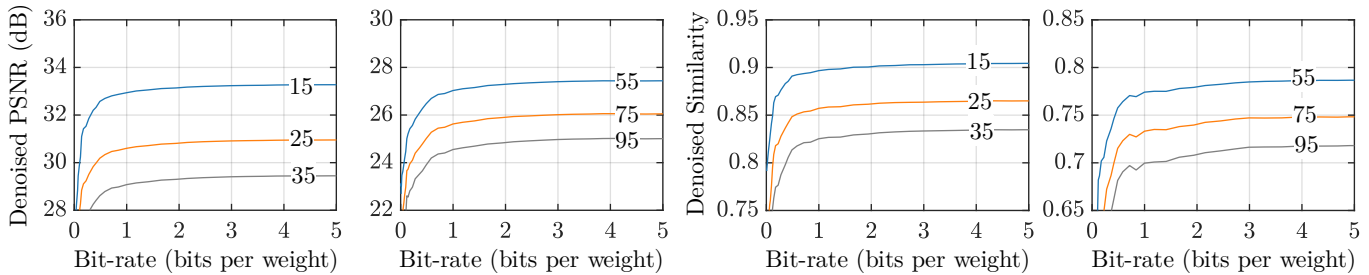

Fig. 10. Average PSNR and SSIM of denoised Set12 images for noise variances 15, 25, 35, 55, 75, and 95, produced by transform-quantized DRUNet [62] at different bit-rates. Denoising performance of DRUNet starts to drop below 2 bits per weight.

TABLE 3
Acceleration of ResNet-50 due to zero-quantization (row-ELT) or pruning (others) of convolution kernels

| Method | Accuracy (%) Top-1 ($\Delta$%) | Top-5 ($\Delta$%) | Bit-rate | FLOP ratio | Compression ratio |
|---|---|---|---|---|---|
| Full-precision | 76.1 (–0.0) | 92.9 (–0.0) | 32.0 | 100.0% | 1.0× |
| SoftFP [8] | 74.6 (–1.5) | 92.1 (–0.8) | 32.0 | 58.2% | 1.7× |
| NISP [19] | 75.2 (–0.9) | – | 32.0 | 56.0% | 1.8× |
| FPGM [9] | 74.8 (–1.3) | 92.3 (–0.6) | 32.0 | 46.5% | 2.2× |
| DCP [14] | 75.0 (–1.1) | 92.3 (–0.6) | 32.0 | **44.2%** | 2.3× |
| GDP [13] | 71.9 (–3.2) | 90.7 (–1.6) | 32.0 | 48.7% | 2.1× |
| GBN-50 [12] | 75.2 (–0.7) | 92.4 (–0.3) | 32.0 | 44.9% | 2.2× |
| GBN-60 [12] | **76.2 (+0.3)** | 92.8 (+0.2) | 32.0 | 59.5% | 1.7× |
| ThiNet-50 [10] | 71.0 (–1.9) | 90.0 (–1.1) | 32.0 | **44.2%** | 2.3× |
| ThiNet-70 [10] | 72.0 (–0.8) | 90.7 (–0.5) | 32.0 | 63.3% | 1.6× |
| Ours + retraining | 72.9 (–3.1) | 91.4 (–1.6) | **1.1** | 65.3% | **29.4×** |
| Ours + retraining | 76.0 (–0.1) | **93.0 (+0.1)** | 2.8 | 88.6% | 11.4× |

TABLE 4
Acceleration of transform-quantized (row-KLT) MobileNet-v2 on Google TPU and MIT Eyeriss

| Methods | Train epochs | Comp ratio | Top-1 Accuracy | Inference rate TPU | Eyeriss |
|---|---|---|---|---|---|
| MobileNet-v2 | | | | | |
| Full-precision | – | – | 71.1 (+0.0) | 1504 (1.00) | 64 (1.00) |
| **HAQ** | 30 | 0.6 | **71.2 (+0.1)** | **2067 (1.37)** | **124 (1.94)** |
| HAQ | 30 | 0.4 | 68.9 (–2.2) | 2197 (1.46) | 128 (2.00) |
| **Ours + retrain** | 30 | **7.3** bits | **71.3 (+0.2)** | **2197 (1.46)** | **127 (1.98)** |
| Ours + retrain | 30 | 7.0 bits | 71.0 (–0.1) | 2207 (1.47) | 128 (2.00) |
| Ours + retrain | 30 | 6.4 bits | 70.9 (–0.2) | 2256 (1.50) | 130 (2.03) |
| Ours + retrain | 30 | 6.1 bits | 67.8 (–3.3) | 2336 (1.55) | 133 (2.08) |
| Ours + retrain | 30 | 5.5 bits | 64.8 (–6.3) | 2407 (1.60) | 136 (2.13) |

TABLE 5
PSNR (dB) of 2–4× upsampled images using transform-quantized EDSR (row-KLT w/o retraining)

| Dataset | | EDSR [63] | Factor [55] | Basis [49] | Basis [49] | 1-bits (ours) | 2-bits (ours) | 3-bits (ours) | 4-bits (ours) |
|---|---|---|---|---|---|---|---|---|---|
| Set5 | 2× | 38.19 | 37.95 | 38.09 | 38.12 | 36.81 | 38.10 | 38.25 | **38.27** |
| | 3× | 34.68 | 34.33 | 34.47 | 34.55 | 32.19 | 34.34 | 34.63 | **34.74** |
| | 4× | 32.48 | 32.05 | 32.29 | 32.39 | 30.31 | 32.28 | 32.51 | **32.60** |
| Set14 | 2× | 33.95 | 33.53 | 33.75 | 33.72 | 32.73 | 33.74 | 33.93 | **34.00** |
| | 3× | 30.53 | 30.31 | 30.41 | 30.46 | 29.09 | 30.39 | 30.55 | **30.63** |
| | 4× | 28.81 | 28.54 | 28.63 | 28.69 | 27.49 | 28.74 | 28.87 | **28.93** |
| B100 | 2× | 32.35 | 32.15 | 32.23 | 32.27 | 31.47 | 32.21 | 32.35 | **32.37** |
| | 3× | 29.26 | 29.08 | 29.15 | 29.18 | 28.19 | 29.13 | 29.26 | **29.32** |
| | 4× | 27.72 | 27.55 | 27.62 | 27.64 | 26.85 | 27.66 | 27.74 | **27.79** |
| Urban100 | 2× | 32.97 | 31.99 | 32.38 | 32.46 | 30.12 | 32.51 | 32.96 | **33.05** |
| | 3× | 28.81 | 28.10 | 28.39 | 28.51 | 26.13 | 28.34 | 28.76 | **28.94** |
| | 4× | 26.65 | 25.98 | 26.25 | 26.36 | 24.67 | 26.38 | 26.65 | **26.81** |
| DIV2K | 2× | 34.60 | 34.60 | 34.77 | 34.84 | 33.43 | 34.79 | 35.03 | **35.08** |
| | 3× | 30.91 | 30.91 | 31.06 | 31.11 | 29.29 | 30.97 | 31.23 | **31.34** |
| | 4× | 28.92 | 28.92 | 29.06 | 29.13 | 27.65 | 29.07 | 29.27 | **29.35** |
| Parameters | | 1180k | 136k | **90k** | 164k | 1180k | 1180k | 1180k | 1180k |
| Compression | | – | 11.5% | 7.6% | 13.9% | **3.1%** | 6.3% | 9.4% | 12.5% |

**Pruning effect of quantization.** We show acceleration due to zero quantization (pruning) of less significant kernels at low bit-rates (1–3 bits). Table 3 shows for ResNet-50 the overall acceleration at bit-rates $R = 1.1$ and 3.1 bits, along with the results from other pruning methods. While our FLOP counts are higher than those of specialized pruning techniques, our FLOPs require much lower bit-depths. Therefore, we may be able to achieve further acceleration if specialized hardware can be designed to facilitate low-bit-depth arithmetic. Fig. 7 provides a per-layer breaks-down of the speedup. Numbers in bold indicate the best result in each column.

**Acceleration on H/W.** We measure speed up from transform quantization on high-performance DNN-targeted hardware accelerators Google TPU [82] and MIT Eyeriss [83], adopting SCALE-sim [84] to simulate the time cycles of MobileNet-v2 quantized using our approach or using HAQ [46]. Note that the convolutional blocks of MobileNet-v2 are already in the 2D transform domain (Section 7) so we optimally assign bit-depths to weights and quantize them via Algorithms 1–2. In both HAQ and our approach, we fetch quantized weights to on-chip memory from off-chip, where they are dequantized and fed into processing units. Off-chip memory is accessed in tandem while convolutions are being performed. In this experiment, we also quantize all intermediate activations in a layer-wise fashion for a fairer comparison with HAQ. Our bit-rates are average quantization bit-depths across weights and activations, assuming the batch size of one. We provide in Table 4, the inference rates (in images per second) of our quantized MobileNet-v2 at different bit-rates, and compare them with those of HAQ. Performance of the two methods are comparable with transform quantization obtaining 2197 and 127 inference rate on TPU and Eyeriss, respectively, with 71.3% top-1 accuracy, while for HAQ they are 2067 and 124 respectively, with 71.2% top-1 accuracy (these are shown in bold). Quantized networks are re-trained on the ImageNet training dataset for 30 epochs.

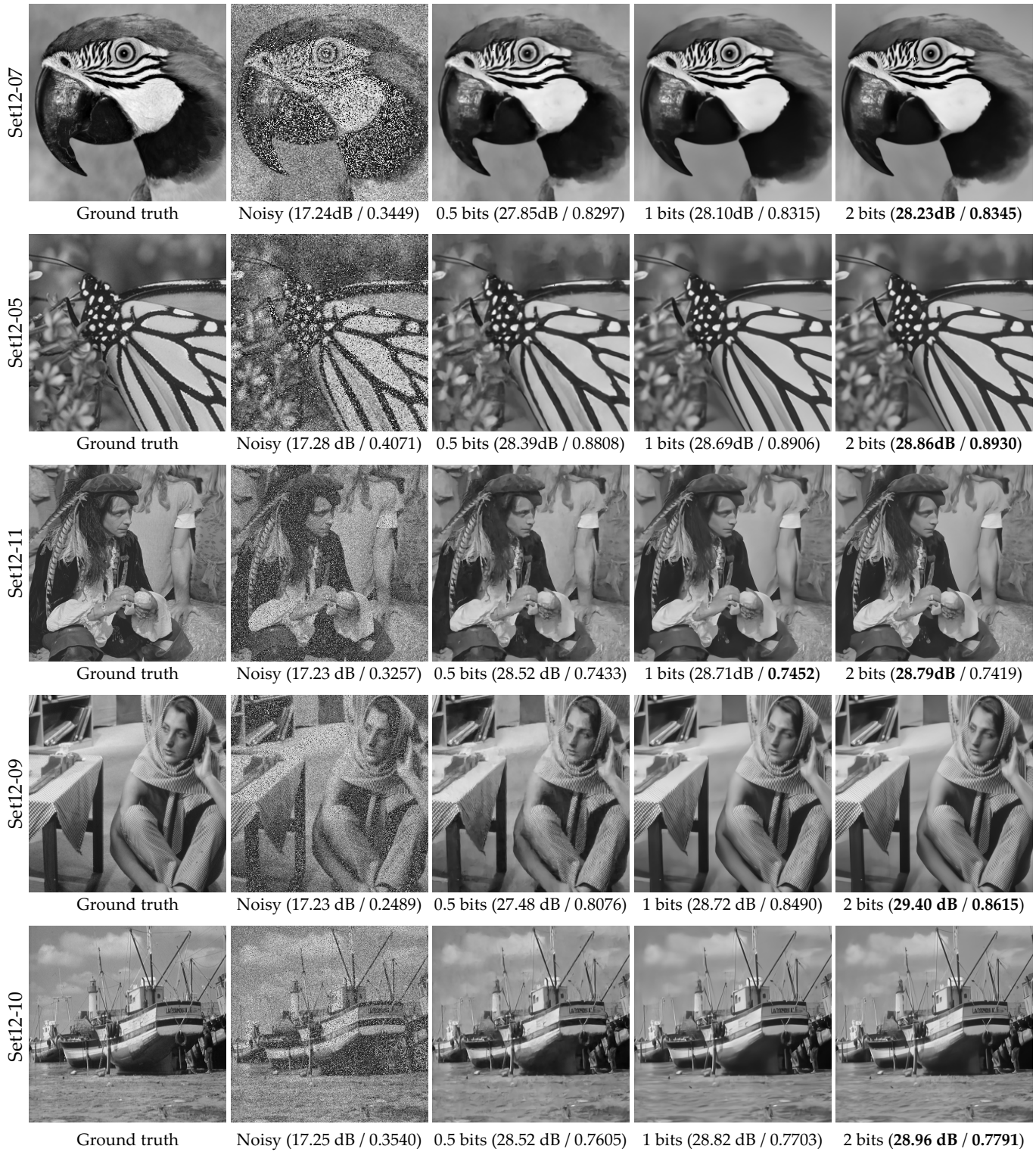

Fig. 11. Visual comparison of denoised Set12 images (noise variance is 35). Denoised images are from row-KLT quantized DRUNet [62] at 0.5, 1.0 and 2.0 bits. PSNR and SSIM in brackets are with respect to ground truth. Images denoised at 2 bits are indistinguishable to the full-precision ones (not shown). Images denoised at 0.5 bits surfer from staircase artifacts. Best viewed online.

**Quantizing image denoising network (DRUNet).** Variants of U-Net have recently been applied to image denoising. We apply row-KLT quantization to the state-of-the-art DRUNet [62] to study the influence of quantization on the denoising performance. We corrupt images by additive Gaussian noise of variance from 15 to 95, and denoise the corrupted images using quantized DRUNets. In Fig. 10, we plot the PSNR and SSIM of the denoised Set12 images against the ground truth across quantization bit-rates and noise variance. We see that PSNR and SSIM of the denoised image start to degrade once quantization rate drops below 2 bits. Fig. 11 visualizes select denoised outputs at the network quantization rates of 0.5, 1.0 and 2.0 bits, together with the noisy input for noise variance 35. We see that KLT-quantized DRUNet (without retraining) has good denoising performance even at a very low bit-rate (0.5 bits). Full precision results are similar to the 2 bit ones.

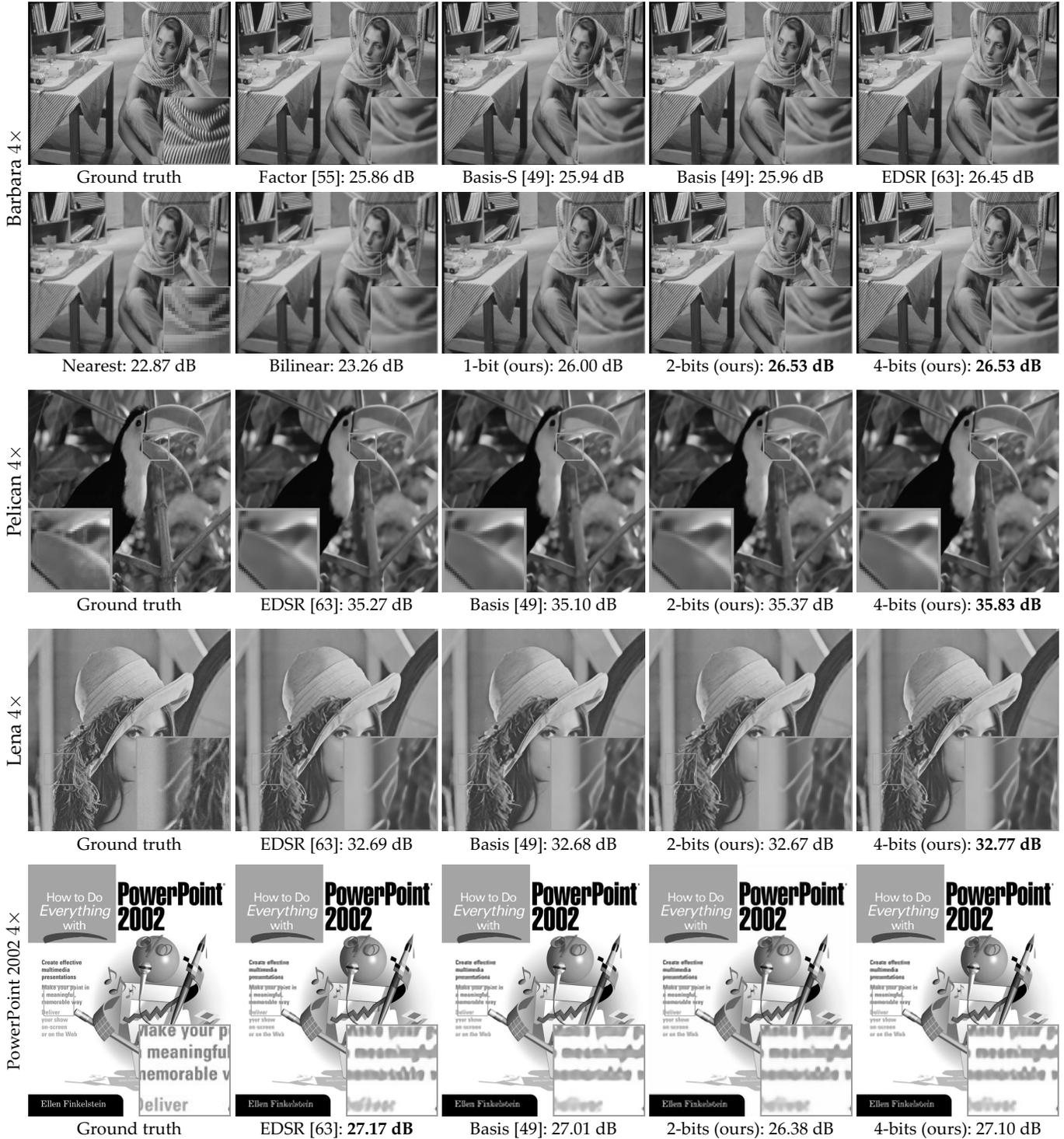

Fig. 12. Visual comparison of 4× upsampled images produced by transform-quantized EDSR [63] (with retraining). We additionally show the upsampled images from Factor [55], Basis [49], and B-spline interpolations for reference. Upsampled images produced by our row-KLT-quantized EDSR at 2–3 bits are visually identical to those of unquantized EDSR. Results for Factor and Basis are taken from [49]. Best viewed online.

**Quantizing super-resolution networks.** We apply row-KLT quantization to EDSR [63] to demonstrate the applicability of our framework to image super-resolution. The pretrained model from [63] is quantized and tested on DIV2K [85], Set5 [86], Set14 [87], B100 [88] and Urban100 [89] datasets. We list the PSNR of the upsampled images in Table 5 together with those from baseline methods, Factor [56], and Basis [49], for comparison. Even without retraining, transform-quantized EDSR outperforms baseline methods (requiring training) in PSNR and compression ratio. We note that our unquantized baseline is little higher (0.1 dB) than those of [49, 56]. Fig. 12 provides a visual comparison among images upsampled by quantized models. Images from transform-quantized EDSR at low bit-rates are very similar to those of the baseline. The bolded numbers indicate the best result in each row.

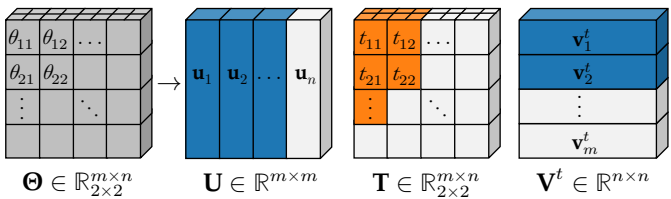

Fig. 13. 2D Transform of $\boldsymbol{\Theta}$. The kernels $\boldsymbol{\Theta}$ are now decorrelated in both the row and the column spaces to produce a sparser matrix $\mathbf{T}$ at lower quantization rates. White blocks indicate zero-quantized regions.

TABLE 6
Intra-kernel transform coding gains (dB) for AlexNet.
Gradient, weight and total (in bold) gains shown.

| $l$ | Intra-kernel transform quantization | | | | | | | | | | | | | | | Inter- | |
| --- | --- | --- | --- | --- | --- | --- | --- | --- | --- | --- | --- | --- | --- | --- | --- | --- | --- |
|  | DCT-I | | | DCT-II | | | KLT | | | ELT | | | | | | KLT | ELT |
| \multicolumn{18}{c}{AlexNet (CaffeNet)} |
| 1 | 14.1 | 5.6 | **19.6** | 14.0 | 5.0 | **19.0** | 11.0 | 5.9 | **16.9** | 14.0 | 5.8 | **19.8** | | | | **4.7** | **6.0** |
| 2 | 4.3 | 2.6 | **6.9** | 4.5 | 1.5 | **5.9** | 3.1 | 3.1 | **6.2** | 4.7 | 2.9 | **7.6** | | | | **6.3** | **7.5** |
| 3 | 2.9 | 2.1 | **5.1** | 3.0 | 1.6 | **4.6** | 2.5 | 2.1 | **4.6** | 3.1 | 2.1 | **5.2** | | | | **3.7** | **5.1** |
| 4 | 2.3 | 1.4 | **3.7** | 2.3 | 1.2 | **3.5** | 2.0 | 1.5 | **3.5** | 2.4 | 1.5 | **3.9** | | | | **1.6** | **3.0** |
| 5 | 2.0 | 2.1 | **4.1** | 2.0 | 1.9 | **3.9** | 1.9 | 2.2 | **4.1** | 2.1 | 2.1 | **4.2** | | | | **1.7** | **3.3** |

## 7 DISCUSSION

Transform quantization may be extended in directions that are not explored in this work. We now discuss each of these extensions, together with possible limitations of our work.

### 7.1 Extension to 2D Transform

In this work, we decorrelated weight matrices $\boldsymbol{\Theta}$ only across their rows. One may wonder if a 2D transform $\mathbf{T} = \mathbf{U}^t\boldsymbol{\Theta}\mathbf{V}$ in both rows and columns of $\boldsymbol{\Theta}$ can provide better compression at lower bit-rates. We illustrate this in Fig. 13. Unfortunately for compression, we find that additional storage overheads incurred from the column transform basis outweigh the bit-savings obtained from a sparser $\mathbf{T}$ (this is especially the case since column transforms do not work as well as row ones— refer to Fig. 6. If the quantized and stored size of the CNNs is not an issue, 2D transforms can still be useful for pruning to reduce the number of convolutions performed. In fact, we can interpret the MobileNet-v2 blocks as having been trained in the 2D transform domain with diagonal $\mathbf{T}$ imposed.

### 7.2 Extension to Intra-kernel Transform

Correlation may also be observed between weights inside a kernel, and one can alternatively transform weight matrices by decorrelating weights within each kernel (intra-kernel) rather than across kernels (inter-kernel). Often, intra-kernel transforms can produce larger coding gains than their inter-kernel counterparts. However, certain CNN models such as AlexNet and VGG can contain significant numbers of $1 \times 1$ convolutional or fully connected layers, so the overall utility of intra-kernel transforms is negligible for such CNNs.

In Table 6, we provide the intra-kernel coding gains of different transforms on the convolutional layers of AlexNet (layers 1–5). All convolutional layers contain kernels $3 \times 3$ in dimensions, with the exception of the first ($11 \times 11$) and the second ($5 \times 5$) layers. For intra-ELT and intra-KLT, we use a separate transform for each row of the weight matrices for improved decorrelation. The coding gains of the intra-ELT are up to 5 dB higher than those of DCT-II used by the CNN quantization methods in [24, 26]. However, the performance of DCT-I is already very similar to that of weight-dependent KLT and ELT, which, unlike the DCT-I, do not have efficient butterfly algorithms available.

### 7.3 Limitations and Future Work

Storage overheads incurred by transform basis matrices can be reduced if the transforms could be shared across multiple layers, similarly to the shared filter basis idea [45, 49]. In our post-training quantization case, however, shared transforms are not useful since the rows or columns of the learnt weight matrices are ordered arbitrarily—transforms that work well on one weight matrix may even yield negative coding gains if applied on different weight matrices. Nevertheless, shared transforms may provide further coding gains if we can train models from scratch to enforce the same order of rows and columns across multiple weight matrices.

One potential limitation of this work is that quantization introduces a significant loss in classification accuracies if we do not subsequently fine-tune the CNN. However, loss due to quantization is far from unique to this work—to the best of our knowledge, CNN quantization methods with smaller accuracy losses involve some form of training, whether it be from scratch or post-quantization. An important theoretical future work may be to investigate the reasons for such a loss in performance when weights are not retrained. Combining quantization with network architecture search similar to [90] may be another fruitful research direction.

## 8 CONCLUSION

This work proposes a transform-quantization framework to compress CNN weights, post-training. As suggested by the name, our framework transforms weights before quantizing them to significantly improve compression in both retrained and non-retrained quantization scenarios. We optimize both components of our framework (transform and quantization) by forming CNN compression as a rate–distortion problem and optimizing its object with respect to the transform, and bit-depths assigned to transform weights and bases. Solving this optimization problem also reveals that the classical KLT has a similar (but not identical) quantization performance as the optimum end-to-end learned transform (ELT) we derive in this paper. For either transform, optimum bit-depths are obtained by minimizing the distortion in the CNN output as opposed to that of the weights themselves. Transform basis layers are quantized similarly to transform domain layers.

The post-training quantization paradigm adopted by our framework also provides the flexibility to compress models at arbitrary bit-rates, obviates the need to train models again from scratch, and is straight-forward to implement. Despite the simplicity of our framework, it advances the state of the art in CNN compression for a number of CNN models. We believe that our transform-quantization framework will also be useful for other CNNs not specifically mentioned here.

# APPENDIX A  PROOFS OF THEOREMS

## A.1  Proof of Theorem 1

*Proof.* If we express $\boldsymbol{\Theta}^q = \boldsymbol{\Theta} + \mathbf{d\Theta}$ for the quantized version of a weight matrix $\boldsymbol{\Theta} \in \mathbb{R}^{n \times m}_{a \times b}$, we have for small $\mathbf{d\Theta}$

$$\begin{aligned}
D &= \tfrac{1}{n}\mathbb{E}\|f(\mathbf{x}|\ldots,\boldsymbol{\Theta}_{l-1},\boldsymbol{\Theta}^q,\boldsymbol{\Theta}_{l+1},\ldots) - \mathbf{y}\|_2^2 \\
&\stackrel{(a)}{\approx} \tfrac{1}{n}\mathbb{E}\|f(\mathbf{x}|\boldsymbol{\Theta}) + \textstyle\sum_{j=1}^m (\partial f(\mathbf{x}|\boldsymbol{\Theta})/\partial \boldsymbol{\theta}_j)^t \mathbf{d\theta}_j - \mathbf{y}\|_2^2 \\
&\stackrel{(b)}{=} \tfrac{1}{n}\textstyle\sum_{j=1}^m \mathbb{E}\|(\partial \mathbf{y}/\partial \boldsymbol{\theta}_j)^t \mathbf{d\theta}_j\|_2^2 \\
&\stackrel{(c)}{=} \tfrac{1}{n}\textstyle\sum_{k=1}^n (\mathbf{C}_{\gamma\gamma})_{kk}(\mathbf{C}_{\theta\theta})_{kk}\epsilon^2 2^{-2R_k} \\
&\stackrel{(d)}{\geq} (\textstyle\prod_{k=1}^n (\mathbf{C}_{\gamma\gamma})_{kk}(\mathbf{C}_{\theta\theta})_{kk})^{1/n} \epsilon^2 2^{-2R}
\end{aligned} \quad (15)$$

in which (a) follows from first-order Taylor's approximation of $f(\mathbf{x}|\ldots,\boldsymbol{\Theta} + \mathbf{d\Theta},\ldots)$ at $\boldsymbol{\Theta}$, (b) follows from the definition $\mathbf{y} = f(\mathbf{x}|\ldots,\boldsymbol{\Theta},\ldots)$ and the fact that gradients $\partial f(\mathbf{x}|\boldsymbol{\Theta})/\partial \boldsymbol{\theta}_j$ are orthogonal. Equality (c) follows from the definitions of the covariance matrices $\mathbf{C}_{\gamma\gamma} = \sum_{j=1}^m (\partial \mathbf{y}/\partial \boldsymbol{\theta}_j)(\partial \mathbf{y}/\partial \boldsymbol{\theta}_j)^t$, and $\mathbf{C}_{\theta\theta} = \tfrac{1}{m}\sum_{j=1}^m \boldsymbol{\theta}_j \boldsymbol{\theta}_j^t$, and the fact that $\mathbb{E}(\mathrm{d}\theta_k)^2 = \mathbb{E}\theta_k^2 2^{-2R_k}$ and $\mathbb{E}\mathrm{d}\theta_{k'}\mathrm{d}\theta_k = 0$, $k' \neq k$. Finally, inequality (d) is obtained with equality for $R = \tfrac{1}{n}\sum_{k=1}^n R_k$ by choosing $R_k$ such that all the individual distortions $(\mathbf{C}_{\gamma\gamma})_{kk}(\mathbf{C}_{\theta\theta})_{kk}\epsilon^2 2^{-2R_k}$ are equal. ∎

## A.2  Proof of Theorem 2

*Proof.* Writing $\mathbf{ST}^q = \mathbf{S}(\mathbf{T} + \mathbf{dT})$ for the quantized version of a weight matrix $\mathbf{ST} \in \mathbb{R}^{n \times m}_{a \times b}$, we have for small $\mathbf{dT}$

$$\begin{aligned}
D &= \tfrac{1}{n}\mathbb{E}\|f(\mathbf{x}|\ldots,\boldsymbol{\Theta}_{l-1},\mathbf{ST}^q,\boldsymbol{\Theta}_{l+1},\ldots) - \mathbf{y}\|_2^2 \\
&\stackrel{(a)}{\approx} \tfrac{1}{n}\mathbb{E}\|f(\mathbf{x}|\boldsymbol{\Theta}) + \textstyle\sum_{j=1}^m (\partial f(\mathbf{x}|\boldsymbol{\Theta})/\partial \boldsymbol{\theta}_j)^t \mathbf{U}^{-t}\mathbf{dt}_j - \mathbf{y}\|_2^2 \\
&\stackrel{(b)}{=} \tfrac{1}{n}\textstyle\sum_{j=1}^m \mathbb{E}\|(\partial \mathbf{y}/\partial \boldsymbol{\theta}_j)^t \mathbf{U}^{-t}\mathbf{dt}_j\|_2^2 \\
&= \tfrac{1}{n}\textstyle\sum_{j=1}^m \mathbb{E}\|(\partial \mathbf{y}/\partial \boldsymbol{\theta}_j)^t \mathbf{U}^{-t}(\mathbf{U}^t \mathbf{d\theta}_j)\|_2^2 \\
&\stackrel{(c)}{=} \tfrac{1}{n}\textstyle\sum_{k=1}^n (\mathbf{U}^{-1}\mathbf{C}_{\gamma\gamma}\mathbf{U}^{-t})_{kk}(\mathbf{U}^t \mathbf{C}_{\theta\theta}\mathbf{U})_{kk}\epsilon^2 2^{-2R_k} \\
&\stackrel{(d)}{\geq} (\textstyle\prod_{k=1}^n (\mathbf{U}^{-1}\mathbf{C}_{\gamma\gamma}\mathbf{U}^{-t})_{kk}(\mathbf{U}^t \mathbf{C}_{\theta\theta}\mathbf{U})_{kk})^{1/n} \epsilon^2 2^{-2R}
\end{aligned} \quad (16)$$

in which (a) follows from first-order Taylor's approximation of $f(\mathbf{x}|\ldots,\mathbf{S}(\mathbf{T} + \mathbf{dT}),\ldots)$ about $\boldsymbol{\Theta} = \mathbf{ST}$, and the chain rule of differentiation

$$(\partial \mathbf{y}/\partial \mathbf{t}_j)^t = (\partial f(\mathbf{x}|\mathbf{ST})/\partial \mathbf{t}_j)^t = (\partial f(\boldsymbol{\Theta})/\partial \boldsymbol{\theta}_j)^t \mathbf{S}, \quad (17)$$

(b) follows from $f(\mathbf{x}|\boldsymbol{\Theta}) = \mathbf{y}$, and (c), from the definitions of covariance matrices $\mathbf{C}_{\gamma\gamma}$, $\mathbf{C}_{\theta\theta}$, and that $\mathbb{E}(\mathrm{d}\theta_k)^2 = \mathbb{E}(\theta_k^2)2^{-2R_k}$ and $\mathbb{E}\mathrm{d}\theta_{k'}\mathrm{d}\theta_k = 0$ for $k' \neq k$. Inequality (d) is obtained with equality for $R = \tfrac{1}{n}\sum_{k=1}^n R_k$ by choosing $R_k$ such that all the distortions $(\mathbf{U}^{-1}\mathbf{C}_{\gamma\gamma}\mathbf{U}^{-t})_{kk}(\mathbf{U}^t\mathbf{C}_{\theta\theta}\mathbf{U})_{kk}\epsilon^2 2^{-2R_k}$ are equal. ∎

## A.3  Proof of Theorem 3

*Proof.* The denominator of (11) further gives us

$$\begin{aligned}
(\sigma_{\mathrm{tc}}^2)^n &= \textstyle\prod_{k=1}^n (\mathbf{U}^{-1}\mathbf{C}_{\gamma\gamma}\mathbf{U}^{-t})_{kk}(\mathbf{U}^t \mathbf{C}_{\theta\theta}\mathbf{U})_{kk} \\
&\stackrel{(a)}{\geq} \det(\mathbf{U}^{-1}\mathbf{C}_{\gamma\gamma}\mathbf{U}^{-t})\det(\mathbf{U}^t \mathbf{C}_{\theta\theta}\mathbf{U}) \\
&= \det(\mathbf{U}^{-1})\det(\mathbf{C}_{\gamma\gamma})\det(\mathbf{C}_{\theta\theta})\det(\mathbf{U}), \\
&= \det(\mathbf{C}_{\gamma\gamma}\mathbf{C}_{\theta\theta}) = \textstyle\prod_{k=1}^n \lambda_k(\mathbf{C}_{\gamma\gamma}\mathbf{C}_{\theta\theta}),
\end{aligned} \quad (18)$$

in which $\lambda_k(\mathbf{A})$ denotes the $k$th eigenvalue of $\mathbf{A}$. Inequality (a) follows from Hadamard's inequality, and all subsequent equalities from the properties of the determinant.

Inequality (a) is obtained with equality if we set $\mathbf{U}$ to the matrix of eigenvectors of $\mathbf{C}_{\gamma\gamma}\mathbf{C}_{\theta\theta}$. Writing $\boldsymbol{\Lambda}$ for the diagonal matrix of eigenvalues $\lambda_k(\mathbf{C}_{\gamma\gamma}\mathbf{C}_{\theta\theta})$, we obtain (12) by writing the eigen-decomposition $\mathbf{U}^{-1}(\mathbf{C}_{\gamma\gamma}\mathbf{C}_{\theta\theta})\mathbf{U} = \boldsymbol{\Lambda}$ as generalized eigenvalue decomposition of matrix pair $(\mathbf{C}_{\theta\theta}, \mathbf{C}_{\gamma\gamma}^{-1})$. ∎

# APPENDIX B  KLT, ELT, SVD, GSVD, HOSVD

To see how the KLT $\mathbf{U}^t$ of $\boldsymbol{\theta} \in \mathbb{R}^{n \times 1}$ relates to the SVD of the corresponding fully-connected weight matrix $\boldsymbol{\Theta} \in \mathbb{R}^{n \times m}_{1 \times 1}$, we notice that cross-channel covariances $\mathbf{C}_{\theta\theta} = (1/m)\boldsymbol{\Theta}\boldsymbol{\Theta}^t$, so the diagonalization $\mathbf{U}^t\mathbf{C}_{\theta\theta}\mathbf{U} = \boldsymbol{\Lambda}$ is equivalent to the SVD

$$\boldsymbol{\Sigma} = \mathbf{U}^t \boldsymbol{\Theta} \mathbf{V}, \quad \mathbf{U}^t \mathbf{U} = \mathbf{I}, \quad \mathbf{V}^t \mathbf{V} = \mathbf{I}, \quad (19)$$

in which $\boldsymbol{\Sigma} = \sqrt{n\boldsymbol{\Lambda}}$, and $\mathbf{V}^t = \boldsymbol{\Sigma}^{-1}\mathbf{U}^t\boldsymbol{\Theta}$. Therefore, the right singular vectors $\mathbf{V}$ of $\boldsymbol{\Theta}$ are transform coefficients of weights (up to scaling by $\boldsymbol{\Sigma}^{-1}$) whereas the left ones $\mathbf{U}$, the transform basis. Similarly, the ELT $\mathbf{U}^t$ of $\boldsymbol{\theta} \in \mathbb{R}^{n \times 1}$ relates to the GSVD (generalized SVD) relation

$$\boldsymbol{\Sigma} = \mathbf{U}^{-1}\boldsymbol{\Theta}\mathbf{V}^{-t}, \quad \mathbf{U}^t\mathbf{C}_{\gamma\gamma}^{-1}\mathbf{U} = \mathbf{I}, \quad \mathbf{V}^t\mathbf{V} = \mathbf{I}, \quad (20)$$

noting that $\mathbf{U}^{-1} \neq \mathbf{U}^t$, and $\mathbf{V}^{-t} \neq \mathbf{V}$ for the GSVD. While the connection between the KLT and the SVD is simple, optimal quantization of $\boldsymbol{\Theta}$ is not immediately obvious in the SVD or GSVD domain, and best understood via the rate–distortion-theoretic framework we have developed in this work.

CNN model compression methods based on high-order-SVD (HOSVD) treat convolution weight matrices as fourth-order tensors, and decompose them across multiple axes (or dimensions). Given their relatively small spatial dimensions (usually no larger than $3 \times 3$), these tensors are decomposed typically only in their input and output-channel dimensions [52]. In this case, one can express the resulting higher-order SVD of the weight tensors equivalently as the 2D transform

$$\mathbf{T} = \mathbf{U}^t \boldsymbol{\Theta} \mathbf{V}, \quad \mathbf{U}^t\mathbf{U} = \mathbf{I}, \quad \mathbf{V}^t\mathbf{V} = \mathbf{I}, \quad (21)$$

in which transforms $\mathbf{U}$ and $\mathbf{V}$ are matrices of eigenvectors of $\mathbf{C}_{\theta\theta} = (1/m)\boldsymbol{\Theta}\boldsymbol{\Theta}^t$, and $\mathbf{C}_{\vartheta\vartheta} = (1/n)\boldsymbol{\Theta}^t\boldsymbol{\Theta}$, respectively, and $\mathbf{T}$ is a non-diagonal matrix of transformed kernels.

**Sean I. Young** received B.Com and B.E. degrees in software engineering from the University of Auckland in 2008, and the M.EngSc and Ph.D. degrees from the University of New South Wales, Sydney, in 2011 and 2018, respectively. He is currently a postdoctoral researcher at Stanford University, Stanford, CA, USA. In 2016, he was a visiting researcher at InterDigital Communications, San Diego, CA. His research interests are large-scale optimization and inverse problems in image processing. He received the APRS/IAPR best paper award at DICTA 2018, together with David Taubman.

**Wang Zhe** received his B.E. degree from Beijing Jiaotong University in 2012, and the M.S. degree in computer applied technology from Peking University in 2015. He is currently a senior research engineer at the Institute for Infocomm Research, A*STAR, Singapore. In 2018, he was a visiting scholar at Stanford University, Stanford, CA, USA. His research interests include large-scale image and video search, neural network compression, and deep learning hardware. He has published more than 20 research papers in computer vision and deep learning, and has filed 5 patents. His work on fast indexing and global descriptor aggregation has been adopted by the MPEG-CDVS (MPEG Compact Descriptors for Visual Search) standard.

**David Taubman** received B.S. and B.E. degrees in electrical engineering from the University of Sydney, in 1986 and 1988, respectively, and the M.S. and Ph.D. degrees from the University of California at Berkeley, in 1992 and 1994, respectively. From 1994 to 1998, he was with Hewlett–Packard's Research Laboratories, Palo Alto, CA. He joined UNSW in 1998, where he is currently a Professor with the School of Electrical Engineering and Telecommunications. He has authored the book JPEG2000: Image Compression Fundamentals, Standards and Practice, with M. Marcellin. His research interests include highly scalable image and video compression, motion estimation and modeling, inverse problems in imaging, perceptual modeling, and multimedia distribution systems. He received the University Medal from the University of Sydney. He has received two best paper awards from the IEEE Circuits and Systems Society for the 1996 paper entitled A Common Framework for Rate and Distortion-Based Scaling of Highly Scalable Compressed Video, and from the IEEE Signal Processing Society for the 2000 paper entitled High Performance Scalable Image Compression with EBCOT.

**Bernd Girod** received the Engineering Doctorate degree from University of Hannover, Germany, and the M.S. degree from Georgia Institute of Technology. Until 1999, he was a Professor with the Electrical Engineering Department, University of Erlangen–Nuremberg. He is currently the Robert L. and Audrey S. Hancock Professor of Electrical Engineering, Stanford University, CA, USA. He has authored over 600 conference and journal papers and six books. His research interests are in the area of image, video and multimedia systems. As an entrepreneur, he was involved in numerous startup ventures, among them Polycom, Vivo Software, 8×8, and RealNetworks. He is a EURASIP Fellow, a member of the National Academy of Engineering, and a member of the German National Academy of Sciences (Leopoldina). He received the EURASIP Signal Processing Best Paper Award in 2002, the IEEE Multimedia Communication Best Paper Award in 2007, the EURASIP Image Communication Best Paper Award in 2008, the EURASIP Signal Processing Most Cited Paper Award in 2008, the EURASIP Technical Achievement Award in 2004, and the Technical Achievement Award of the IEEE Signal Processing Society in 2011.